\documentclass[10pt,journal,compsoc]{IEEEtran}

\usepackage{booktabs}       % professional-quality tables
\usepackage{nicefrac}       % compact symbols for 1/2, etc.
\usepackage{microtype}      % microtypography
\usepackage[x11names]{xcolor}         % colors
\usepackage{algorithm}
\usepackage{algorithmic}
\usepackage{amsmath,amsfonts}
\usepackage[hidelinks]{hyperref}
\usepackage{mathtools}
\usepackage{bbding}
\usepackage{multirow}
\usepackage{subfigure}
\usepackage{colortbl}
\usepackage{caption}
\usepackage{amsthm}
\theoremstyle{plain}
\newtheorem{theorem}{Theorem}[section]

\theoremstyle{definition}
\newtheorem{definition}[theorem]{Definition}

\theoremstyle{remark}

\ifCLASSOPTIONcompsoc
  \usepackage[nocompress]{cite}
\else
  \usepackage{cite}
\fi

\ifCLASSINFOpdf
\else
\fi

\begin{document}
\title{Individual and Structural Graph Information
Bottlenecks for Out-of-Distribution Generalization}
% Ling Yang
% ~Ling_Yang1
% , Jiayi Zheng, Heyuan Wang, Yinghong Li, Shenda Hong
% \author{Ling~Yang,
%         Jiayi Zheng, Heyuan Wang, Zhilin Huang, Yinghong Li, \\Shenda Hong, Wentao Zhang, and Bin Cui
\author{Ling~Yang, Jiayi Zheng, Heyuan Wang, Zhongyi Liu, \\Zhilin Huang, Shenda Hong, Wentao Zhang, and Bin Cui, \IEEEmembership{Senior Member,~IEEE}
% \author{Ling~Yang, Zhilin Huang, Shenda Hong, Wentao Zhang, and Bin Cui, \IEEEmembership{Senior Member,~IEEE}
% \IEEEcompsocthanksitem L. Yang, Z. Huang, S. Hong, and B. Cui are with Peking University, Beijing, China. Email: yangling0818@163.com, zhengjiayi980@126.com, heyuanww@163.com, \{zerinhwang03, hongshenda, bin.cui\}@pku.edu.cn.
\IEEEcompsocitemizethanks{
% \IEEEcompsocthanksitem  is with the Institute of Medical Technology, Health Science Center of Peking University, Beijing, 100191, China. Email: yangling@stu.pku.edu.cn.
\IEEEcompsocthanksitem L. Yang, J. Zheng, H. Wang, Z. Huang, and S. Hong are with Peking University, Beijing 100871, China. Email: yangling0818@163.com, \{zheng.jiayi, hy.wang, zerinhwang03, hongshenda\}@pku.edu.cn. Z. Liu is with Ant Group, Beijing, China. Email: zhongyi.lzy@antgroup.com.
\IEEEcompsocthanksitem B. Cui is with School of CS \& Key Lab of High Confidence Software Technologies (MOE), Institute of Computational Social Science, Peking University (Qingdao), Beijing, China. Email: bin.cui@pku.edu.cn. W. Zhang is with Mila - Québec AI Institute, HEC Montréal, Canada. Email: wentao.zhang@mila.quebec.
\IEEEcompsocthanksitem L. Yang and J. Zheng contributed equally to this work. W. Zhang and B. Cui are corresponding authors.

\IEEEcompsocthanksitem This work was supported by the National Natural Science
Foundation of China (No.62102008, No.61832001, and U22B2037).
}

}

\markboth{IEEE TRANSACTIONS ON KNOWLEDGE AND DATA ENGINEERING}%
{Shell \MakeLowercase{\textit{et al.}}: A Sample Article Using IEEEtran.cls for IEEE Journals}


\IEEEtitleabstractindextext{%
\begin{abstract}
Out-of-distribution (OOD) graph generalization are critical for many real-world applications. Existing methods neglect to discard spurious or noisy features of inputs, which are irrelevant to the label. Besides, they mainly conduct instance-level class-invariant graph learning and fail to utilize the structural class relationships between graph instances. 
In this work, we endeavor to address these issues in a unified framework, dubbed \textbf{I}ndividual and \textbf{S}tructural \textbf{G}raph \textbf{I}nformation \textbf{B}ottlenecks (\textit{IS-GIB}).
To remove class spurious feature caused by distribution shifts, we propose Individual Graph Information Bottleneck (I-GIB) which discards irrelevant information by minimizing the mutual information between the input graph and its embeddings.
To leverage the structural intra- and inter-domain correlations, we propose Structural Graph Information Bottleneck (S-GIB). Specifically for a batch of graphs with multiple domains, S-GIB first computes the pair-wise input-input, embedding-embedding, and label-label correlations. Then it minimizes the mutual information between input graph and embedding pairs while maximizing the mutual information between embedding and label pairs. The critical insight of S-GIB is to simultaneously discard spurious features and learn invariant features from a high-order perspective by maintaining class relationships under multiple distributional shifts. Notably, we unify the proposed I-GIB and S-GIB to form our complementary framework IS-GIB. Extensive experiments conducted on both node- and graph-level tasks consistently demonstrate the superior generalization ability of IS-GIB. The code is available at https://github.com/YangLing0818/GraphOOD.
\end{abstract}

}

\maketitle

\IEEEdisplaynontitleabstractindextext

% \fi
\IEEEpeerreviewmaketitle

\section{Introduction}
Graph-structured data is now ubiquitous in the real-world applications, such as biomedical networks \cite{wu2018moleculenet}, social networks \cite{easley2012networks}, recommender systems \cite{wu2020graph} and knowledge graphs \cite{wang2017knowledge}. Recent graph neural networks \cite{kipf2016semi,velivckovic2017graph,xu2018powerful,wu2019simplifying,yang2020dpgn,fan2020graph,yu2021self,li2021higher,miao2021lasagne,yang2022omni,cheng2020strict,shu2020host,shu2022multi,yang2022diffusionscene,tang2019coherence,zhan2018graph,zhan2017graph} have increasingly emerged as dominant approaches for processing the graphs. However, existing GNNs are mostly based on the assumption that the testing and training graph data are independently sampled from the identical distribution, i.e., the I.I.D. assumption. Therefore, they are sensitive to the domain of the dataset and may exhibit significant performance degradation when encountering distribution shifts between training and testing data \cite{shen2021towards,li2022out}. 

Actually, there exist a wide range of possible complex distributional shifts in realistic graph-structured data involving the changes of feature, size and structure \cite{ding2021closer,ji2022drugood}. 
For instance, in drug discovery, the causal effect between scaffolds (the input) and their property (the label) would be invalid with different test distributions of molecules \cite{hu2020open}.
For social networks, the input and the label denote the distributions of users’
friendships and activities respectively, which highly depend on when/where the networks are collected \cite{fakhraei2015collective,qiu2018deepinf}. 
In financial networks \cite{pareja2020evolvegcn}, the flows of payment between transactions (the input) and the appearance of illicit transactions (the label) may be closely correlated with some external contextual factors, such as time and market. 
Therefore, it is vital to learn GNNs capable of out-of-distribution generalization, and thus enable to achieve relatively stable performances under distribution shifts, especially for some high-risk applications, e.g., medical diagnosis \cite{li2020graph,yang2022unsupervised},
criminal judicature \cite{han2020risk}, financial business \cite{yang2019using}, and drug discovery \cite{wu2018moleculenet}, etc.

Despite the works on domain generalization and OOD generalization have achieved great success \cite{arjovsky2019invariant,sagawa2019distributionally,ahuja2021invariance}, few efforts are made to develop the OOD graph generalization \cite{li2022ood,bevilacqua2021size}. 
A series of effective methods is  to conduct domain-invariant graph learning \cite{chen2022invariance,wu2021discovering}.
% which exploits the invariant relationships between features and labels across different distributions by disregarding the variant spurious correlations. 
On the one hand, some methods encourage the similarity of graph representations across multiple
environments to learn environment-invariant representations \cite{zhu2021shift,yehudai2021local}.
% Specifically, \citet{zhu2021shift} proposes to address node-level OOD generalization in GNNs by explicitly minimizing the distributional differences between biased training data and true inference distribution of graphs. \citet{yehudai2021local} demonstrates the importance of local structure and proposes a self-supervised method aimed at learning meaningful representations of local structures that appear in large graphs.
On the other hand, some methods are built upon the invariance principle \cite{wu2021discovering} to address the OOD graph generalization problem from a more principle way \cite{wu2021handling}.
% Specifically, \citet{wu2021handling} proposes a guaranteed solution for the node-level OOD problem by minimizing the mean and variance of risks from multiple training environments that are generated by the environment generators. DIR \cite{wu2021discovering} is proposed to handle
% graph-level OOD generalization tasks by discovering invariant rationales for GNN. 

However, existing domain-invariant graph learning methods have two limitations: 1) they only focus on learning invariant feature highly relevant to the label and overlook the importance of explicit restriction on discarding spurious feature about categorial information, 2) they mainly conduct instance-level invariant graph learning and fail to utilize the structural class relationships between graph instances, which are vital for overall classification performance.

In this work, we address the above limitations in a unified framework. Inspired by the Information Bottleneck (IB) theory \cite{tishby2000information},  we propose a first-order Individual  Graph Information Bottleneck (I-GIB) to address the first issue. It adds instance-wise restriction on each graph to discard irrelevant information by minimizing the mutual information between the input graph and its representations, and simultaneously learns invariant features capture all the information about the label by maximizing the mutual information between the representations and corresponding labels. To address the second issue, we propose a second-order Structural Graph Information Bottleneck (S-GIB) to add pair-wise restrictions within intra- and inter-domain graphs. Specifically for a batch of graphs with multiple domains, S-GIB first computes the pair-wise input-input, embedding-embedding, and label-label correlations. Then it minimizes the mutual information between input and embedding pairs while maximizing the mutual information between embedding and label pairs. The critical insight of S-GIB is to simultaneously discard spurious features and learn invariant features from a structural or high-order perspective by maintaining class relationships under multiple distributional shifts on graphs. Notably, we unify the proposed first- and second-order information bottlenecks (I-GIB and S-GIB) to
form a complementary framework IS-GIB. Extensive experimental result on various datasets with different OOD scenarios proves our IS-GIB outperforms other baseline by a large margin. To summarize, our paper makes the following contributions:
\begin{itemize}
    \item We firstly propose Individual Graph Information Bottlenecks (I-GIB) to impose explicit instance-wise constraints for discarding spurious feature in node- or graph-level representations for OOD graph generalization.
    \item To the best of our knowledge, we firstly propose Structural Graph Information Bottlenecks (S-GIB) to exploit pair-wise intra- and inter-domain correlations between node or graph instances, which encourages learning graph invariant feature from a high-order perspective.
    \item We combine I-GIB and S-GIB in a complementary framework, dubbed IS-GIB, to conduct graph invariant learning. And we theoretically derive a tractable optimization objective for IS-GIB in practice.
    \item We conduct comprehensive experiments on various datasets of both node- and graph-level classification tasks in different scenarios of distribution shifts, proving that our framework can mitigate distribution shift better than existing methods.
\end{itemize}

\section{Related Work}
\label{related work}
\textbf{Information Bottleneck.}
The Information Bottleneck (IB) principle \cite{tishby2000information} is originally proposed for data compression and signal processing in the field of information theory. VIB \cite{alemi2016deep} firstly bridges the gap between IB and the deep learning. In last few years, IB has been widely employed in many fileds due to the capability of learning compact and meaningful representations, such as computer vision \cite{luo2019significance,zhang2021coarse,li2021invariant}, natural language processing \cite{wang2020learning,mahabadi2021variational} and reinforcement learning \cite{igl2019generalization,liu2021learning}. Nevertheless, existing IB-based methods mainly focus on instance-level optimization \cite{goldfeld2020information,yu2020graph,wu2020graph2}, and thus fail to address more structural and complex problems such as OOD graph generalization.\\

\noindent \textbf{Invariant Learning for Out-of-Distribution Graph Generalization.}
\label{RW-1}
Invariant learning \cite{arjovsky2019invariant,chang2020invariant,ahuja2021invariance} aims to exploit the invariant relationships between features and labels across different distributions while disregarding the variant spurious correlations.
Recent works utilize the invariant learning in OOD graph generalization problem and can provably achieve satisfactory performance under distribution shifts \cite{zhu2021shift,zhang2021stable}. The most critical part of them is how to design invariant learning tasks and add proper regularization specified for extracting domain-invariant representations \cite{wu2021discovering,wu2021handling,yehudai2021local}. Besides, some works \cite{bevilacqua2021size,chen2022invariance} begin to combine the causal model \cite{sugiyama2007covariate} to further promote the invariant learning. However, they only concentrate on learning invariant feature highly relevant to the label, and thus overlook the importance of adding explicit restrictions to discard spurious features or correlations, such as the structural relationships within intra- and inter-domain graphs, i.e, class relationships.\\ 
% \subsection{Imbalance Problem in Graph Neural Networks}
% Imbalanced classification problems \cite{he2009learning} are widespread in real scenarios including graph-based applications \cite{shi2020multi}. Most existing methods focus on class-imbalance problem \cite{zhao2021graphsmote,ghorbani2022ra} while \citet{chen2021topology} emphasize the importance of topology-imbalance problem. In OOD graph generalization, the  

\section{Preliminaries}
\subsection{Node- and Graph-level OOD Generalization}
We first introduce the problem definitions of node- and graph-level OOD generalization in detail. 
%In this paper, we address both OOD graph generalization tasks in an identical way since node-level task is handled with 
With respect to graph-level task, let $\mathbf{G}_{tr} = \{\mathcal{G}_i\}_{i=1}^{N}$ and $\mathbf{G}_{te} = \{\mathcal{G}_i\}_{i=1}^{M}$ 
be the training and testing graph datasets, which are under distribution shifts, i.e., $p(\mathbf{G}_{tr}) \neq p(\mathbf{G}_{te})$ and $\mathbf{G}_{te}$ is unobserved in the training stage. The goal is to train a generalizing-to-OOD GNN model $f_\theta=(f_{enc}, f_{cls})$, which consists of a graph encoder $f_{enc}$ and a graph classifier $f_{cls}$. The graph encoder $f_{enc}:\mathcal{G}\rightarrow \mathcal{Z}$ maps the input graph space $\mathcal{G}$ to a d-dimensional representation space $\mathcal{Z}\in \mathbb{R}^d$. The graph classifier $f_{cls}:\mathcal{Z}\rightarrow \mathcal{Y}$ maps the representation space $\mathcal{Z}$ to the label space $\mathcal{Y}\in \mathbb{R}^c$, where $c$ denotes the number of classes. The definition of node-level task is similar to the graph-level task, the main difference is that the node-level task use ego-graphs centered at target nodes as the inputs. In this paper, we formulate both tasks in a same way for simplicity. As illustrated \cite{gilmer2017neural}, the $l$-th layer of a GNN model can be formulated as:
\begin{equation}
    h_v^{(l)} = f^{(l)}_{U} \big( h_v^{(l-1)} , f^{(l)}_{M} \big( \big\{ \big( h_v^{(l-1)}, h_u^{(l-1)}, X_{uv} \big) : u \in \mathcal{N}(v) \big\} \big) \big),
\end{equation}
where $\mathcal{N}(v)$ is the neighborhood set of $v$, $h_v^{(l)}$ denotes the representation of node $v$ at the $l$-th layer, $h_v^{(0)}$ is initialized as the node attribute $X_v$, and $f^{(l)}_{M}$ and $f^{(l)}_{U}$ stand for the message passing and update function at the $l$-th layer respectively. Since $h_v$ summarizes the information of a subgraph centered around node $v$, the entire graph's embedding can be derived as below:
\begin{equation} \label{eq2}
h_{\mathcal{G}} = f_{R} \big( \big\{ h_{v} | v \in \mathcal{V} \big\} \big) ,
\end{equation}
where $f_{R}$ is a permutation-invariant readout function, \emph{e.g.} mean pooling or more complex graph-level pooling function \cite{ying2018hierarchical}. Particularly, $f_{enc}$ comprises $f_{U}$, $f_{M}$ and $f_{R}$. 
\subsection{Graph Invariant Learning}
\label{gil-intro}
As illustrated in Section.\ref{RW-1}, graph invariant learning is a main stream of OOD graph generalization tasks. Here we provide formulations of existing methods. Some methods adopt regularizers
to encourage the similarity of representations across multiple environments to learn environment-invariant representations, which can be formulated as:
\begin{equation}
\small
\min_{f_\theta} \mathbb{E}_{\mathcal{G}, \mathcal{Y}} [\ell(f_\theta(\mathcal{G}), \mathcal{Y})] + \ell_{reg},
\end{equation}
where $\ell_{reg}$ denotes the loss of the designed regularizer. Another group of methods treat the cause of distribution shifts between testing and training graph data as a potential unknown environmental variable $e$. The optimization objective can be formulated as:
\begin{equation}
\min_{f_\theta} \max_{e\in\mathcal{E}} \mathcal{R}(f_\theta|e) = \min_{f_\theta} \max_{e\in\mathcal{E}} \mathbb{E}^e_{\mathcal{G}, \mathcal{Y}} [\ell(f_\theta(\mathcal{G}), \mathcal{Y})],
\end{equation}
where $\mathcal{E}$ is the support of training environments and $\mathcal{R}(f_\theta|e)$ is the risk of the $f_\theta$ on the training environment $e$. Next, we will elaborate our proposed Individual and Structural
Information Bottlenecks respectively, and illustrate how we combine both complementary information bottlenecks in a unified graph invariant learning. 

\section{Methodology}
\label{sec-method}
In this section, we elaborate our Individual and Structural Graph Information Bottlenecks (IS-GIB) for OOD graph Generalization, which consists of I-GIB and S-GIB. We first introduce the proposed Individual Graph Information Bottleneck (I-GIB) in Section.\ref{sec-i-gib}. Then we introduce the proposed Structural Graph Information Bottleneck (S-GIB) in Section.\ref{sec-s-gib}. Finally, we derive a tractable optimization objective
for our IS-GIB in Section.\ref{sec-is-gib}, and summarize the whole pipeline of proposed framework.
\begin{figure*}[h]
\centering
    \includegraphics[width=11cm]{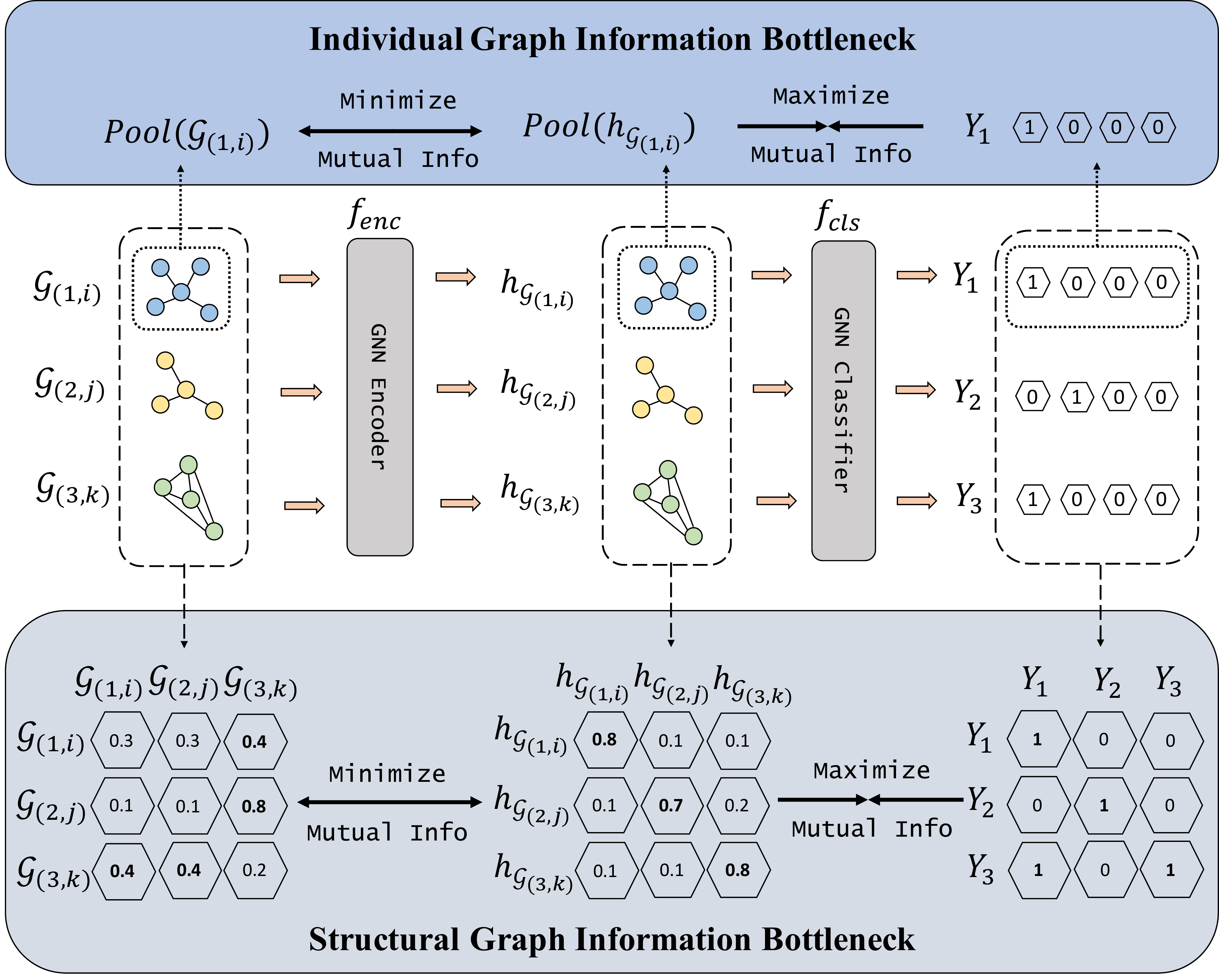}
\caption{Illustration of Individual and Structural Graph Information Bottlenecks.}
\label{fig:method}
% \vspace{-6mm}
\end{figure*}
\subsection{Individual  Graph Information Bottleneck}
\label{sec-i-gib}
\textbf{Motivations.} Regarding the instance-level graph invariant learning, existing methods have proposed some effective objectives to find informative substructures or subgraphs for OOD graph-related tasks \cite{yu2020graph,yu2022improving,miao2022interpretable}. However, they can not adequately remove the possible spurious features or noises since they mainly make efforts in learning domain-invariant features for OOD graph generalization tasks.
An intuitive way is to additionally restrict on the spurious features, but it is difficult to define the spurious features especially in latent space of intermediate layers.
Inspired from Information Bottleneck theory, we propose the Individual Graph Information Bottleneck (I-GIB) to produce an efficient information flow for processing each graph as an alternative way, which imposes a constraint between input and extracted representations to enable minimal sufficient information to pass though the embedding network. In this way, the obtained representations will be more relevant to the corresponding class by get ridding of irrelevant information.
\begin{definition}[\textbf{Individual Graph Information Bottleneck}] Let $\mathcal{G}_{(i,j)}$ be the $i$-th graph with the $j$-th distribution shift in the training batch, given a graph GNN model $f_\theta=(f_{enc},f_{cls})$ and corresponding label $Y_i$, the I-GIB seeks for
the most informative yet compressed graph representation $h_{\mathcal{G}_{(i,j)}}$ by optimizing the following objective:
\begin{equation}
\begin{aligned}
\max_{f_{enc}}{I(Y_i,h_{\mathcal{G}_{(i,j)}})}  \quad \text{ s.t. } I(\mathcal{G}_{(i,j)},h_{\mathcal{G}_{(i,j)}}) \leq I_{c}.
\end{aligned}
\label{eq3}
\end{equation}
where $I_{c}$ is the information constraint between $\mathcal{G}_{(i,j)}$ and $h_{\mathcal{G}_{(i,j)}}$. By introducing a Lagrange multiplier $\beta$ to
Eq.\ref{eq3}, we reach its unconstrained form:
\begin{equation}
\begin{aligned}
\max_{f_{enc}}{I(Y_i,h_{\mathcal{G}_{(i,j)}})}-\beta I(\mathcal{G}_{(i,j)},h_{\mathcal{G}_{(i,j)}}).
\end{aligned}
\label{eq4}
\end{equation}
\end{definition}
I-GIB employs a constraint between the raw graph and representations while maximizing the mutual information between the graph representations and label.
With the refined information flow,
I-GIB is capable of discarding spurious feature and properly retains sufficient invariant feature highly relevant to the label. 
However, the I-GIB objective in Eq.\ref{eq4} is notoriously hard to optimize due to the intractability of mutual information. We will introduce approaches on how to optimize such objective in Section.\ref{sec-is-gib}.
\subsection{Structural Graph Information Bottleneck}
\label{sec-s-gib}
\textbf{Motivations.} In OOD graph generalization tasks, the spurious features within intra- and inter-domain graphs can badly affect the learning process of invariant feature. 
Besides, the variant combinations of distributional shifts or sophisticated environments lead to a hard GNN optimization. 
Therefore, it is critical to learn from a more structural perspective and extract the most informative feature by explicitly utilizing the cross-domain consistencies or relationships.    
Nevertheless, existing graph invariant learning methods mainly focus on instance-level optimization, they all fail to make efforts in leveraging structural relationships. EERM \cite{wu2022towards} demonstrates that it is effective to utilize the topological context in domain-invariant node classification. However, it can not be extend to graph-level tasks since it is based on ego-graph structures. Besides, it also neglects to utilize pair-wise class relationships within graphs, which is more structural and informative for node- and graph-level classification tasks. 
Thus we firstly propose a second-order objective, dubbed Structural Graph Information Bottleneck (S-GIB), to make an adequate use of structural relationships, i.e., class relationships.
\begin{definition}[\textbf{Structural Graph Information Bottleneck}] Given an environment consisting of $N$ graphs with $D$ distributional shifts, a graph GNN model $f_{enc}=(f_{enc},f_{cls})$. Let $\mathcal{G}_{(i,j)}$ be the $i$-th graph with the $j$-th distribution shift, and $Y_{i}$ denotes the corresponding label, the S-GIB structurally seeks for a set of the most informative yet compressed graph representations $\{h_{\mathcal{G}_{(i,j)}}\}$ that maintain the class relationships (\textbf{we omit the superscript and subscript of the sets $\{...\}_{i=1}^N$ for simplicity}):
\begin{equation}
\begin{aligned}
&\max_{f_{enc}}{I(\mathcal{R}(\{{Y}_{i}\}),\mathcal{R}(\{h_{\mathcal{G}_{(i,j)}}\}))},  \\ \text{ s.t. } &I(\mathcal{R}(\{\mathcal{G}_{(i,j)}\}),\mathcal{R}(\{h_{\mathcal{G}_{(i,j)}}\})) \leq I_{c}.
\end{aligned}
\label{eq5}
\end{equation}
where $R(\cdot)$ is the deterministic pair-wise relationships-computing function (inner product by default, more discussions in Section.\ref{ablation-relationship-mea}), take the example of labels relationships:
\begin{equation}
\begin{aligned}
\mathcal{R}(\{{Y}_{i}\}) = \begin{pmatrix}
Y_0Y_0 & Y_0Y_1 & \cdots & Y_0Y_{N-1} \\
Y_1Y_0 & Y_1Y_1 & \cdots & Y_1Y_{N-1}  \\
\vdots  & \vdots  & \ddots & \vdots  \\
Y_{N-1}Y_0 & Y_{N-1}Y_1 & \cdots & Y_{N-1}Y_{N-1} 
\end{pmatrix}.
\end{aligned}
\label{eq6}
\end{equation}
% Particularly, we directly use a pooling operation on the inputs for relationships computation. 
Similar to Eq.\ref{eq4}, we reach the unconstrained form of Eq.\ref{eq5}:
\begin{equation}
\begin{aligned}
\max_{f_{enc}}{I(\mathcal{R}(\{{Y}_{i}\}),\mathcal{R}(\{h_{\mathcal{G}_{(i,j)}}\}))}-\beta I(\mathcal{R}(\{\mathcal{G}_{(i,j)}\}),\mathcal{R}(\{h_{\mathcal{G}_{(i,j)}}\})).
\end{aligned}
\label{eq7}
\end{equation}
\end{definition}
S-GIB takes a batch of cross-domain graphs, and computes pair-wise relationships (second-order information) within inputs, representations and labels.
Particularly, we directly use a pooling operation on the inputs for relationships computation.
Then it minimizes the mutual information between inputs and representations correlations to discard spurious connections, and simultaneously maximizing the mutual information between representations and labels correlations to strengthen invariant connections. Through this formulation, S-GIB is able to learn invariant feature that only relates to the class information from a more structural and global perspective since the relationships between graphs with variant distributional shifts should be also invariant from the input to the prediction. To efficiently discard spurious feature and capture invariant feature in OOD graph generalization tasks, we incorporate S-GIB with I-GIB in an integral framework to conduct graph invariant learning. Nevertheless, both I-GIB and S-GIB objectives are hard to optimize due to the intractability of mutual information. Next, we will elaborate approaches on how to optimize both objectives. 
\subsection{Optimizing Integral Graph Invariant Learning}
\label{sec-is-gib}
As illustrated above, the I-GIB and S-GIB enhance the OOD graph generalization ability from instance- and structure-level respectively. 
To simultaneously leverage both powerful objectives, we combine them to conduct integral graph invariant learning. 
Before that, we first explore the optimization of each objective to solve the intractability of mutual information.

Motivated by Variational Information Bottleneck (VIB) \cite{alemi2016deep}, we use variational approximation to help the optimization. Given $R_H=\mathcal{R}(\{h_{\mathcal{G}_{(i,j)}}\})$, $R_{\mathcal{G}}=\mathcal{R}(\{\mathcal{G}_{(i,j)}\})$ and $R_Y=\mathcal{R}(\{{Y}_{i}\})$, we rewrite the Eq.\ref{eq7} as:
\begin{equation}
\begin{aligned}
\small
{I(R_Y,R_{H}))}-\beta I(R_{\mathcal{G}},R_H) \geq \mathbb{E}_{p{(R_Y, R_H)}}\Big[ \log q_\phi(R_Y|R_H)\Big]  \\-  \beta \mathbb{E}_{p(R_{\mathcal{G}}),p(R_H)}\Big[ \log \frac{p(R_H|R_{\mathcal{G}})}{r(R_H)}\Big].
\end{aligned}
\label{eq8}
\end{equation}
where $q_\phi$ is a MLP projection layer and we obtain a lower bound. However, the prior distribution $r(R_H)$ is still hard to estimate. Therefore, we borrow the idea from \cite{donsker1975asymptotic} about the KL-divergence, and we obtain an alternative form of the lower bound:
\begin{equation}
\begin{aligned}
\small
&\mathbb{E}_{p{(R_Y, R_H)}}\Big[ \log q_\phi(R_Y|R_H)\Big]+\beta\Big[\mathbb{E}_{p{(R_{\mathcal{G}}, R_H)}}\Big[f_{\phi}(R_{\mathcal{G}},R_H)\Big] \\&-\mathbb{E}_{p(R_{\mathcal{G}}),\ p(R_H)}\Big[\log e^{f_{\phi}(R_{\mathcal{G}},R_H)}\Big]\Big].
\end{aligned}
\label{eq9}
\end{equation}
where $f_{\phi}$ consists of a MLP-based projection layer and distance function (i.e., $\ell_2$ distance).

In Eq.\ref{eq8} and Eq.\ref{eq9}, we provide a lower bound for the proposed S-GIB, and here we provide more details about how to yield them. For the first term in Eq.\ref{eq8}, we expand it as:
\begin{equation}
\begin{aligned}
I(R_Y,R_H)& = \int p(R_Y,R_H) \log{{p(R_Y|R_H)}} dR_Y\ dR_H \\ &\ \ \ \ - \int p(R_Y,R_H) \log{{p(R_Y)}} dR_Y\ dR_H  \\
& = \int p(R_Y,R_H) \log{{p(R_Y|R_H)}} dR_Y \ dR_H  + \mathrm{H}(R_Y).
\end{aligned}
\label{eq-proof1}
\end{equation}
where $\mathrm{H}(R_Y)$ is the entropy of $R_Y$ and thus can be ignored in later optimization. However, optimizing Eq.\ref{eq-proof1} is still a difficult task. Hence, we approximate $p(R_Y,R_H)$ with an empirical distribution $p(R_Y,R_H)\approx\frac{1}{N}\sum_{i=1}^{N}\delta_{R_Y}(R_Y^i)\delta_{R_H}(R_H^i)$, where $\delta()$ is the Dirac function to sampling training data. where $R_H$ and $R_Y$ are the pair-wise correlations of the graph embeddings and labels respectively. Next, we substitute the true posterior $p(R_Y|R_H)$ with a variational approximation $q_{\phi_{1}}(R_Y|R_H)$ as \cite{alemi2016deep}, we obtain a tractable lower bound of the first term in Eq.\ref{eq8}:
\begin{equation}
\begin{aligned}
I(R_Y,R_H)& \geq \int p(R_Y,R_H) \log{{q_{\phi}(R_Y|R_H)}} dR_Y \ dR_H \\ &\ \ \ \ \ + \mathrm{KL}(p(R_Y|R_H)|q_{\phi}(R_Y|R_H)) \\
&\geq \int p(R_Y,R_H) \log{{q_{\phi}(R_Y|R_H)}} dR_Y \ dR_H \\
&\approx \frac{1}{N} \sum_{i=1}^{N} q_{\phi}(R_Y^{i}|R_H^i).
\end{aligned}
\end{equation}
The second term in Eq.\ref{eq8} directly borrows the idea from \cite{donsker1975asymptotic} about the KL-divergence. Similarly, the proposed I-GIB also has its tractable lower bound as formulated in Eq.\ref{eq11}. 

Considering a $N$-graphs batch, we randomly generate the virtual environment $e_i$ containing the graph $\mathcal{G}_{(i,j)}$. With this setting, we can combine I-GIB with S-GIB in batch training.
We combine Eq.\ref{eq7}, Eq.\ref{eq8} and Eq.\ref{eq9}, use $f_{cls}$ to substitute the $q_\phi$ and formulate the tractable optimization objective of S-GIB as:
\begin{equation}
\begin{aligned}
% \mathcal{L}_{S-GIB}&=\ \mathcal{L}_{S,1}+\mathcal{L}_{S,2}\\&=\
&\mathcal{L}_{S-GIB}=\mathcal{L}_{S,1}+\mathcal{L}_{S,2}=\\
&\frac{1}{N}\sum_{e_i=1}^{N} \mathcal{L}_{cls}\Big(\mathcal{R}(\{{Y}_{i}\}), \mathcal{R}(\{f_{cls}({h_{\mathcal{G}_{(i,j)}})}\})|e_i\Big)\\&-\beta\Big[\frac{1}{N}\sum_{e_i=1}^{N}f_{\phi}\Big(\mathcal{R}(\{\mathcal{G}_{(i,j)}\}),\mathcal{R}(\{h_{\mathcal{G}_{(i,j)}}\})|e_i\Big) \\&-\log \frac{1}{N}\sum_{e_i=1,e_k\neq e_i}^{N} \exp\Big({f_{\phi}(\mathcal{R}(\{\mathcal{G}_{(k,m)}\}|e_k),\mathcal{R}(\{h_{\mathcal{G}_{(i,j)}}\}|e_i))}\Big)\Big].
\end{aligned}
\label{eq10}
\end{equation}
where the mutual information maximization is denoted by $\mathcal{L}_{S,1}$, and the mutual information restriction with coefficient $\beta$ is denoted by $\mathcal{L}_{S,2}$. $m$ and $j$ are both the indices of distribution shifts, $\mathcal{L}_{cls}$ is the classification loss. Similar to S-GIB, we obtain the tractable optimization objective of I-GIB as:
\begin{equation}
\begin{aligned}
&\mathcal{L}_{I-GIB}= \mathcal{L}_{I,1}+\mathcal{L}_{I,2}=\\
&\frac{1}{N}\sum_{i=1}^{N} \mathcal{L}_{cls}\Big(Y_i,f_{cls}( h_{\mathcal{G}_{(i,j)}})\Big) -\beta \Big[ \frac{1}{N}\sum_{i=1}^{N}f_{\phi}\Big(\mathcal{G}_{(i,j)},h_{\mathcal{G}_{(i,j)}}\Big) \\
 &-\log \frac{1}{N}\sum_{i=1,k\neq i}^{N} \exp\Big({f_{\phi}(\mathcal{G}_{(k,m)},h_{\mathcal{G}_{(i,j)}})}\Big)\Big].
\end{aligned}
\label{eq11}
\end{equation}
Finally we get the total optimization objective by weighted sum of $\mathcal{L}_{I-GIB}$ and $\mathcal{L}_{S-GIB}$ as:
\begin{equation}
\begin{aligned}
 \mathcal{L}_{total}=\mathcal{L}_{I,1}+\gamma_1\mathcal{L}_{I,2}+\gamma_2\mathcal{L}_{S,1}+\gamma_3\mathcal{L}_{S,2},
\end{aligned}
\label{eq12}
\end{equation}
where the coefficients $\gamma_1,\gamma_2,\gamma_3$ will be further discussed in latter experimental section.
% For a better illustration of our IS-GIB, we take the graph-level OOD generalization task as the example and summarize our training procedure in Algorithm.\ref{alg}. 
% \begin{algorithm}
% \caption{Optimizing the Individual and Structural Graph Information Bottlenecks.}
% \label{alg}
% \begin{algorithmic}
%   \STATE {\bfseries Input:} Graph dataset $\mathbf{G} = \{\mathcal{G}\}_{i=1}^{N}$, graph label set $\mathbf{Y} = \{Y_i\}_{i=1}^{N}$, the number of training steps $T$. 
%   \STATE {\bfseries Output:} A domain-invariant GNN model $f_{\theta_T}$.
% %   \STATE Pre-train the GNN with local self-supervised method and 
%   \STATE Initialize the model parameters $f_{\theta_0}$.
%     \STATE Initialize the empty environment set $\mathcal{E}=\{\}$.
%   \FOR{$t=1$ {\bfseries to} $T$}

%   \STATE Sample a mini-batch of graphs $\{\mathcal{G}_i\}_{i=1}^M \in \mathbf{G}$.
%   \FOR{$i=1$ {\bfseries to} $M$}
%   \STATE Randomly generate the environment $e_i$ consists of $K$ graphs across domains.
%   \STATE Enqueue the $e_i$ to the environment set $\mathcal{E}$.
%   \ENDFOR
%   \STATE Obtain I-GIB objective $\mathcal{L}_{I-GIB}$ with $\mathbf{G}=\{\mathcal{G}_i\}_{i=1}^M$ according to Eq.\ref{eq11}. 
%   \STATE Obtain S-GIB objective $\mathcal{L}_{S-GIB}$  with $\mathcal{E}=\{e_i\}_{i=1}^M$ according to Eq.\ref{eq10}.
%   \STATE Update the model parameters $f_{\theta_t}$ according to Eq.\ref{eq12}.
%   \STATE Empty the $\mathcal{E}$.
%  \ENDFOR
% \end{algorithmic}
% \end{algorithm}
\subsection{Comparison with Existing Methods}
\textbf{Information Bottleneck on GNN-Related Taks.} Some existing works \cite{yu2020graph,yu2022improving,miao2022interpretable} have utilized IB principle \cite{tishby2000information} in GNN-related researches, and prove its effectiveness. They all aim to find or predict useful and informative subgraphs in graphs for classification \cite{yu2020graph} and interpretation \cite{yu2022improving,miao2022interpretable} tasks. Nevertheless, such utilization is at instance level, it may fail to handle the complex structural OOD environments. In contrast, our IS-GIB first proposes a structural GIB, and unify it with the proposed individual GIB to tackle the OOD problem.\\  

\noindent \textbf{OOD Generalization on Graphs.} Recent works begin to handle the OOD generalization problems on graphs from different perspectives \cite{bevilacqua2021size,wu2021discovering,wu2021handling,chen2022invariance}. DIR \cite{wu2021discovering} and GOOD \cite{chen2022invariance} both learn to recognize the invariant subgraphs to make OOD graph generalization. \cite{bevilacqua2021size} uses a causal model to make better size extrapolation between train and test data.
OOD-GNN \cite{li2022ood} employs a novel nonlinear graph representation decorrelation method to remove the spurious correlations. EERM \cite{wu2021handling} designs multiple context explorers to extrapolate from a single observed environment \cite{krueger2021out}. However, they all focus on instance-level invariant graph learning by adapting the model to designed complex environments, and fail to utilize the abundant structural or high-order relationships
within intra- and inter-domain graphs. In contrast, our IS-GIB first explicitly utilize the structural relationships, i.e., class relationships, to optimize and improve the invariant graph learning. Besides, our IS-GIB first imposes explicit instance-wise
constraints for discarding spurious feature in node-
or graph-level representation extraction procedure.

\section{Experiments}
\label{sec-experimentss}
\begin{center}
\begin{table*}[h]\small
\centering
\caption{Summary of Datasets.}
\label{table_datasets_info}
  \setlength{\tabcolsep}{1mm}{
  \begin{tabular}{c|cccccccc}
  \toprule
      {Dataset} &\textsc{Cora} & \textsc{Citeseer}  && \textsc{Twitch-Explicit} & \textsc{Facebook100}  && \textsc{Gossipcop} & \textsc{Politifact}  \\
      \midrule
    %   Scenario & \multicolumn{2}{c}{Artificial Shifts} && \multicolumn{2}{c}{Multi-domain generalization} &&  \multicolumn{2}{c}{Graph-level Generalization} \\
    \vspace{0.2cm}
      Task & \multicolumn{2}{c}{Node classification} && \multicolumn{2}{c}{Node classification} && 
      \multicolumn{2}{c}{Graph classification} \\
      
      \#Nodes & 2708 & 3327 & & 1912 - 9498 & 769 - 41536  && 3 - 199 & 3 - 497 \\
      \#Edges &10556 & 9228 && 31299 - 153138 & 16656 - 1590655 && 2 - 198 & 2 - 496 \\
      \#Classes &7 & 6 && 2 & 2 && 2 & 2 \\
      \#Graphs & 1 & 1 && 7 & 100 && 5464 & 314 \\
      Metric & Accuracy & Accuracy && ROC-AUC &Accuracy &&Accuracy &Accuracy\\
   
    \bottomrule
  \end{tabular}  
  }
%   \vspace{-0.5cm}
  \end{table*}
\end{center}
\textbf{Experimental Settings.} In this section, to verify the effectiveness of our proposed IS-GIB, we conduct extensive experiments on a wide range of public datasets. Specifically, four datasets are from node-level prediction tasks and two target at graph-level prediction tasks.
We consider three different scenarios of distribution shift on graphs: (1) Artificial distribution shifts. (2) Cross-domain generalization. (3) Distribution shifts on graph-level tasks. We compare our method with standard Empirical Risk Minimization (\textbf{ERM}) and two type of solutions for OOD: (1) Robust Optimization including \textbf{GroupDRO} \cite{sagawa2019distributionally}, \textbf{SR-GNN} \cite{zhu2021shift}, \textbf{OOD-GNN}\cite{li2022ood}.  (2) Invariant and causal learning, including \textbf{IRM} \cite{arjovsky2019invariant}, \textbf{GOOD}\cite{chen2022invariance}, \textbf{EERM}\cite{wu2022towards}, \textbf{DIR} \cite{wu2021discovering} and 
Size-Invariant representation(\textbf{Size-inv}) \cite{bevilacqua2021size}.
We also adopt 5 different GNN backbones: \textbf{GCN} \cite{kipf2016semi, chen2020simple}, \textbf{GraphSAGE} \cite{hamilton2017inductive}, Graph Isomorphism Network (\textbf{GIN}) \cite{xu2018powerful}, Graph Attention Network (\textbf{GAT}) \cite{velivckovic2017graph}, Chebyshev Spectral Graph Convolution (\textbf{ChebNet}) \cite{defferrard2016convolutional}. In the following subsections, we discuss the experimental results in different scenarios.

Table \ref{table_datasets_info} describes the details of these  datasests. Recall that we conduct experiments on both node- and graph-level OOD graph generalization tasks. \textbf{Node level}: (1) for artificial distribution shifts, we choose two node classification datasets \textsc{Cora} and \textsc{Citeseer}. (2) Cross-domain generalization means for multiple graphs with different distributions, we select some of them for training and then test directly on unseen graphs. \textsc{Twitch-Explicit} and \textsc{Facebook100} are two social network datasets with multiple social graphs, the task of these two datasets are still node classification. \textbf{Graph level}: we choose two datasets of fake news detection (graph classification), \textsc{Gossipcop} and \textsc{Politifact}, collected by \cite{dou2021user}. 

\noindent\textbf{Implementation Details.}
\label{sec:imple_detail}
We present the implementation details. We run all of the experiment on Nivida GeForce RTX 2080Ti 11G, the runtime environment is Ubuntu 16.04 LTS with CUDA 11.4, PYTHON=3.8.12, PyTorch=1.8.1, DGL=0.8.0.
In the section \ref{sec_4_2}, we use different GNN backbones, for all of them, we use the implementation of DGL: \textsl{GCN2Conv}, \textsl{GATConv}, \textsl{SAGEConv}, \textsl{GINConv}, \textsl{ChebConv}.
In the section \ref{sec_ablation}, to achieve lower computational complexity, we sample a fixed batch of pairs to compute the pairwise relationships $R(\cdot)$ in Eq. \ref{eq10} (also Eq. \ref{eq7}). Concretely, for node-level classification, we sample a batch of nodes from each graph with size of \emph{b}. Then we gather all the nodes and compute the pairwise relationships between them, as described in \ref{eq6}.
By default, we use \textsl{SAGEConv} as the backbone and set the layer number \emph{L} = 3, the hidden dimension \emph{H} = 64, dropout rate \emph{d} = 1.

For \textsc{Cora} and \textsc{Citeseer}, we tune the following parameters around the following default values: learning rate \emph{lr}={1e-3}; batch for computing pairwise relationships \emph{b}=128 (per graph); weight decay=1e-6; $\gamma_1$=0.5; $\gamma_2$=0.1; $\gamma_3$=0.5.
For \textsc{Twitch-Explicit} and \textsc{Facebook100}, the default values is as follow: learning rate \emph{lr}={1e-4}; batch for computing pairwise relationships \emph{b}=128 (per graph);  weight decay=1e-5; $\gamma_1$=0.75; $\gamma_2$=0.1; $\gamma_3$=1.
For \textsc{Gossipcop} and \textsc{Politifact}, the default values is as follow: learning rate \emph{lr}={1e-3}; batch for computing pairwise relationships \emph{b}=128 (per graph); weight decay=1e-5; $\gamma_1$=0.5; $\gamma_2$=0.05; $\gamma_3$=1.
For all the methods and datasets, we train the model by running a fixed number of epochs and report the test result of the model that achieves the best performance on validation set.

\subsection{Node-level OOD Graph Generalization}
\subsubsection{Handling Artificial Distribution Shifts}
\label{sec_4_1}

% \begin{figure}[htbp]
%     \centering
% 	%\qquad
% % 	\hfill
% 	\begin{minipage}{0.48\linewidth}
% 		\centering\small
% 		\captionof{table}{Test accuracy on \textsc{Cora} and \textsc{Citeseer}, we run 10 times for each algorithm and report the mean value and standard deviation.}
%         \begin{tabular}{ccc}
%         \toprule
%     \textbf{Methods}     & Cora     & Citeseer \\
%     \midrule
%     ERM &  82.73 $\pm$ 2.73 & 88.74 $\pm$ 1.27 \\
%     GroupDRO &  83.45 $\pm$ 3.45 & 87.94 $\pm$ 2.85 \\
%     SR-GNN & 84.76 $\pm$ 2.98  & 89.47 $\pm$ 1.56 \\
%     DIR &   84.12 $\pm$ 3.95 & 89.56 $\pm$ 2.01 \\
%     EERM &  85.01 $\pm$ 3.12 &91.12 $\pm$ 1.45  \\
%     \midrule
%      \textbf{IS-GIB} &  \textbf{89.15 $\pm$ 3.02} & \textbf{93.90 $\pm$ 1.87} \\
%     \bottomrule
%     \end{tabular}
    
%     \label{table_4_1_1}
%     \end{minipage}
% 	    \begin{minipage}{0.42\linewidth}
% 		\centering
% % 		\vspace{-0.6cm}
% 		\setlength{\abovecaptionskip}{0.2cm}
% 		\includegraphics[width=\linewidth]{expr_plot/0501_plot1.pdf}
% 		\caption{Test accuracy of 5 runs on \textsc{Cora} with different noises. ($\mu/\sigma$: mean value/standard deviation)}
% 		\label{fig:1}
% 	\end{minipage}
% 	\vspace{-10pt}
% \end{figure}

First, we consider how to handle the synthetic noises injected to the graph on two public node classification datasets \textsc{Cora} and \textsc{Citeseer}. Both of these two datasets only contain one grpah, to perform artificial distribution shifts, we add extra noise into the node's feature so that we can generate multiple graphs. Specifically, given an encoded feature matrix $A$ with the shape $(n \times d)$, $n$ is the number of nodes and $d$ is the dimension of features. We randomly generate a noise matrix $B$ and add it to the input of GNN, the final input is $A+B$. The noise matrix $B$ follows a normal distribution with a pre-specified mean value $\mu$ and standard deviation $\sigma$. In this paper, we generate 10 graphs and use 4/2/4 graphs for training/validation/testing. In the experiment of the comparison with baselines, for training graphs, we did not add noises; for validation/test graphs, we set $\mu=0, \sigma=1$. The other experiment conducted in this section shows the model's performance in different distribution shift scales, we set $\mu=0, \sigma=1$ for training graphs and select different values of $\mu$ and $\sigma$ for validation/test graphs as described in the following.
\begin{table}[ht]
\begin{center}
\begin{small}
\caption{Test accuracy on \textsc{Cora} and \textsc{Citeseer}}

% \vskip 0.15in
  \label{table_4_1_1}
  \center
  \begin{tabular}{ccc}
        \toprule
    \textbf{Methods}     & Cora     & Citeseer \\
    \midrule
    ERM &  82.73 $\pm$ 2.73 & 88.74 $\pm$ 1.27 \\
    \midrule
    GroupDRO &  83.45 $\pm$ 3.45 & 87.94 $\pm$ 2.85 \\
    SR-GNN & 85.76 $\pm$ 2.98  & 90.47 $\pm$ 1.56 \\
    OOD-GNN & 84.97$ \pm$ 2.45  & 90.69 $\pm$ 1.73 \\
    \midrule
    IRM & 83.65$\pm$ 3.01  & 89.45 $\pm$ 2.08 \\
    GOOD & 84.95 $\pm$ 2.87  & 90.45 $\pm$ 1.56 \\
    Size-inv & 86.23 $\pm$ 1.67  & 90.89 $\pm$ 2.15 \\
    DIR &   84.12 $\pm$ 3.95 & 89.56 $\pm$ 2.01 \\
    EERM &  85.98 $\pm$ 3.12 &91.12 $\pm$ 1.45  \\
    \midrule
     \textbf{IS-GIB} &  \textbf{89.15 $\pm$ 3.02} & \textbf{93.90 $\pm$ 1.87} \\
    \bottomrule
    \end{tabular}
\end{small}
\end{center}
% \vspace{-6mm}
\end{table}
The test results of different OOD generalization methods are shown in Table \ref{table_4_1_1}. We find traditional ERM performed wrost among all the methods.
% these two datasets are small. If do not design sophisticated methods for tackling different noises across domains, it is easy to cause the model to overfit the training domains. 
% We find IRM and SR-GNN perform slightly better than GroupDRO, indicating that there is no good enough method based on meta-learning for tackling distribution shift on graphs.
GroupDRO is a general approach for OOD genralization which is not specifically designed for graph data, and we can observe that in general its performance is inferior to the other robust optimization method specially designed for GNN (SR-GNN, OOD-GNN). Recent studies mainly focus on invariant learning and causal learning, both the causal model and the ideas of extrapolation are useful, and we can see Size-inv, DIR and EERM perform better than other baselines.  
EERM consistently achieves the best result among all of the baselines, while minimizing the variance of risks between different environments, it also exploited adversarial training for exploring more environments. 
But all of them still neglected the fine-grained relationships between different domains and classes. 
Our IS-GIB outperforms all of the baselines.
The improvement of IS-GIB over ERM is 7.76\% and 5.81\% on \textsc{Cora} and \textsc{Citeseer} respectively. 
The objective of existing methods such as V-REx (Variance-Risk-Extrapolation) \cite{krueger2021out} can not adequately remove the noises. By introducing I-GIB, we explicitly discard spurious feature and maintain useful invariant feature for prediction. In addition, the S-GIB considers both intra- and inter-domain structural consistency thereby discarding the structural-level noises.\\

\begin{figure}[h]
%  \vspace{-0.2cm}  %调整图片与上文的垂直距离
 \setlength{\abovecaptionskip}{0cm}   %调整图片标题与图距离
% \setlength{\belowcaptionskip}{-0.1cm}   %调整图片标题与下文距离
% \vspace{-0.5em}
\centering
   \includegraphics[width=0.65\linewidth]{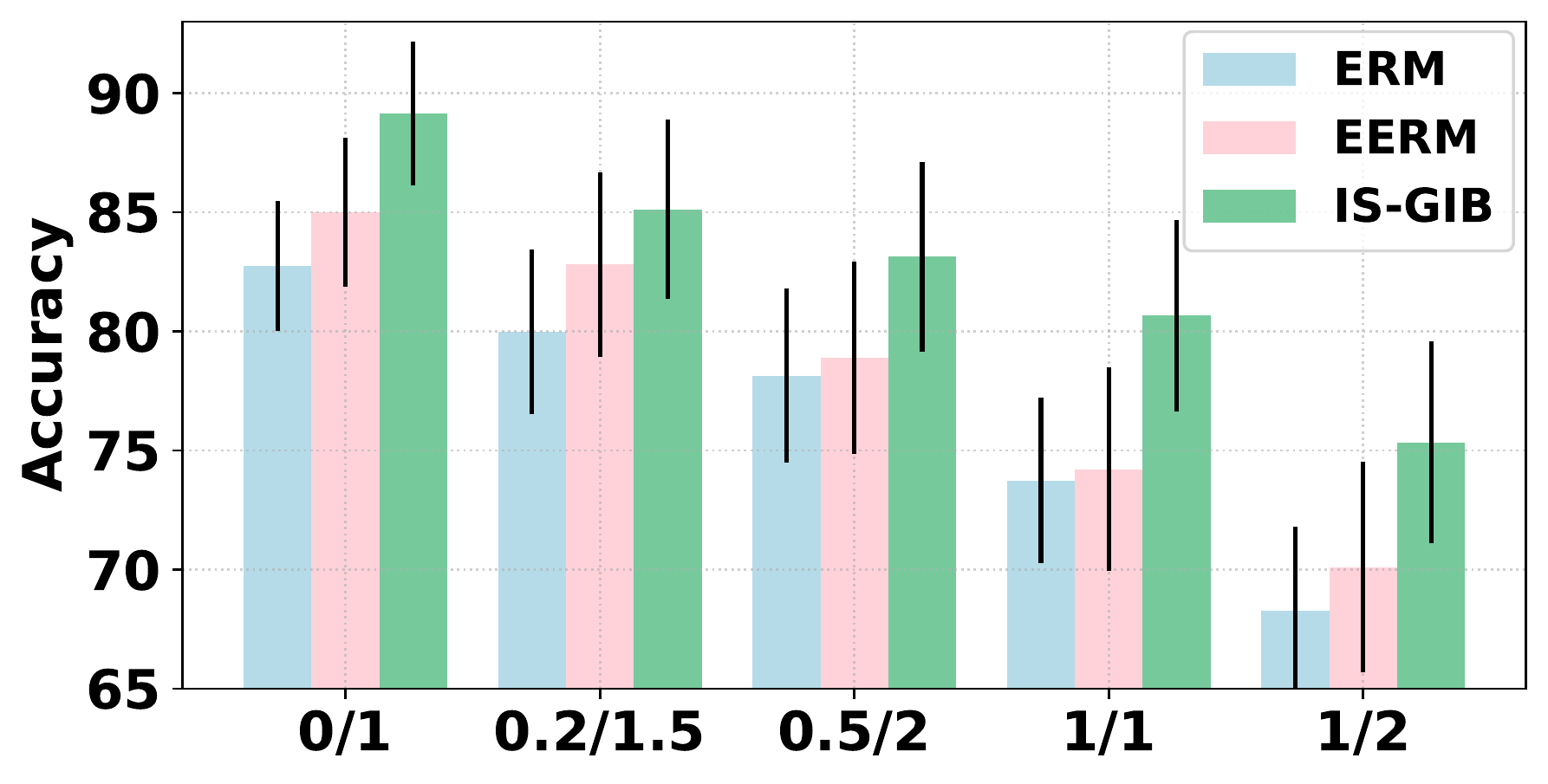}

\caption{Test accuracy of 5 runs on \textsc{Cora} with different noises. ($\mu/\sigma$: mean value/standard deviation)}
\label{fig:1}
\end{figure}
\noindent \textbf{Robustness to the Noises of Variant Scales.} In Figure \ref{fig:1}, we show the test results of 5 different runs with different distribution of artificial noises injected into the validation/test graphs on \textsc{Cora}.
% We add batch normalization operation to each layer of the graph convolution. 
Intuitively, a larger mean value or standard deviation implies more uncertainty and deviation of the noises in different graphs, which means the task is harder.
We set the mean value and standard deviation ($\mu/\sigma$) of these 5 distributions to 0/1,  0.2/1.5, 0.5/2, 1/1, 1/2 respectively.
As expected, more biased noise has greater negative impact on the performance. Intriguingly when the noise is gradually amplified, the performance gap between IS-GIB and other methods also gradually increases. It indicates that IS-GIB can greatly resist the negative impact of noise. We attribute the superior performance to two unique designs of the model: (1) the second item in both I-GIB (Eq. \ref{eq4}) and S-GIB (Eq. \ref{eq7}) directly constrains the propagation of noisy input signals. (2) the S-GIB helps to better distinguish the useful features across domains from a global pair-wise perspective.
 
\begin{table*}[ht]
\begin{center}
\begin{small}
\caption{Performance comparison of test graphs on \textsc{Twitch-Explicit} (ROC-AUC score) and \textsc{Facebook100} (accuracy).}
% \vskip 0.15in
  \label{table_4_2_1_cross_d}
  \center
  \setlength{\tabcolsep}{1.6mm}{
  \begin{tabular}{c|ccccccccc}
  \toprule
\multirow{2}{*}{Methods}  &
\multirow{2}{*}{\space}&
\multicolumn{3}{c}{Twitch-Explicit} &
\multicolumn{1}{c}{\space}& \multicolumn{3}{c}{Facebook100}&
\\ \cmidrule(r){3-5} \cmidrule(r){7-9} 
% \cline{9-10} 
&&T1 & T2 & T3&&T4& T5 & T6  \\
\midrule
ERM&&56.14$\pm$0.90&56.30$\pm$0.59&56.55$\pm$0.99&&51.53$\pm$1.05 &52.26$\pm$0.67&51.03$\pm$0.25 \\
\midrule
GroupDRO&&56.45$\pm$1.33&55.93$\pm$1.01&55.98$\pm$1.28&& 51.02$\pm$0.87&53.04$\pm$1.02&52.29$\pm$1.09\\
SR-GNN&&56.98$\pm$0.56&56.37$\pm$0.25&57.25$\pm$1.03&&52.41$\pm$0.56 & 54.27$\pm$0.99& 53.04$\pm$0.33 \\
OOD-GNN &&55.86$\pm$0.78&56.23$\pm$0.67&56.46$\pm$0.82&&51.98$\pm$0.45 & 54.24$\pm$0.34& 52.60$\pm$0.37 \\
\midrule
IRM &&56.34$\pm$0.67&56.21$\pm$0.36&56.78$\pm$0.95&&52.03$\pm$0.44 & 53.15$\pm$0.64& 51.85$\pm$0.71 \\
GOOD &&56.98$\pm$0.44&56.84$\pm$0.92&57.09$\pm$0.33&&52.27$\pm$0.92 & 53.86$\pm$0.71& 52.64$\pm$0.59 \\
Size-inv &&57.45$\pm$0.66&57.35$\pm$0.47&58.04$\pm$0.83&&52.56$\pm$0.89 & 54.01$\pm$0.91& 53.78$\pm$0.64 \\
DIR&&57.27$\pm$1.76&56.40$\pm$0.76&57.07$\pm$0.56&&52.01$\pm$1.32&53.94$\pm$1.01&53.33$\pm$0.46\\
EERM&&57.32$\pm$0.48&57.01$\pm$0.94&57.95$\pm$1.05&& 53.02$\pm$0.39&53.65$\pm$0.90&54.47$\pm$0.86 \\
\midrule
\textbf{IS-GIB}&&\textbf{58.28$\pm$0.38}&\textbf{58.61$\pm$0.70}&\textbf{59.16$\pm$0.94}&&\textbf{53.63$\pm$0.95}&\textbf{55.78$\pm$0.76}&\textbf{55.24$\pm$0.74}\\
\bottomrule
\end{tabular}}
\end{small}
\end{center}
\vspace{-6mm}
\end{table*}

\subsubsection{Cross-Domain Generalization}
\label{sec_4_2}

We use \textsc{Twitch-Explicit} and \textsc{Facebook100} (\cite{lim2021new}) in cross-domain scenarios, in which the GNN model is trained on multiple graphs and evaluated on the unseen test graphs. 
% In cross-domain generalization scenario, we use two social network datasets \textsc{Twitch-Explicit} and \textsc{Facebook100} \cite{lim2021new}. 
\textsc{Twitch-Explicit} contains seven distinct graphs: DE, ENGB, ES, FR, PTBR, RU and TW, each of them are collected from the corresponding region. The nodes represent the users and edges represent their relationships. \textsc{Facebook100} is also a social network dataset which is collected in 2005. 
In this paper, we use a subset of the whole \textsc{Facebook100}.
To demonstrate the effectiveness of our method, we run multiple times with different combinations of training/validation/test graphs. Concretely, for \textsc{Twitch-Explicit}, we use ENGB for validation, and use \{DE, ES, FR\} for training in T1, use \{DE, ES\} for training in T2, use \{DE, FR\} for training in T3. By default, we use other graphs for testing respectively. In \textsc{Facebook100}, we use \{Brown11, BC17, MU78, NYU9, UC64\} for testing, \{Tufts18, UCSD34, UVA16\} for validation, and we use \{Amherst41, Virginia63, Wake73\} for training in T4, \{Vassar85, Tulane29, Syracuse56, Temple83\} for training in T5, \{UCLA26, Stanford3, Rutgers89, Simmons81\}  for training in T6. 
% We run multiple times with different training/validation/test graph splits, see Appendix.\ref{appendix-cd} for more details about experimental settings.
From the performance comparison of Table \ref{table_4_2_1_cross_d}, we find 
% We can see that IS-GIB outperforms the ERM baseline by a large margin, with a maximum lead of about \textcolor{red}{20\%}. 
IS-GIB consistently performs better than all the baselines.
% proving the second order S-GIB is useful in multi-graph generalization. 
% The performance gap is significant on both the sparse graphs (\textcolor{red}{A, B}, with average density of 0.0) and dense graphs (\textcolor{red}{A, B}, with average density of 0.0).
On \textsc{Facebook100}, compared with ERM, our IS-GIB gains an improvement with up to 8.25\%, which is more significant than that on \textsc{Twitch-Explicit}.Since we use more graphs for training on
\textsc{Facebook100} (11 graphs in total) than in \textsc{Twitch-Explicit} (7 graphs in total), which implies IS-GIB can efficiently catch the intra-domain similarities for a better generalization with the help of structural information bottleneck.
\\

\begin{figure}[ht]
  \subfigure[]{
  
  \begin{minipage}[t]{0.48\linewidth}
\makeatletter\def\@captype{figure}
    \centering
    \label{fig:42_1_1}
    \includegraphics[scale=0.32]{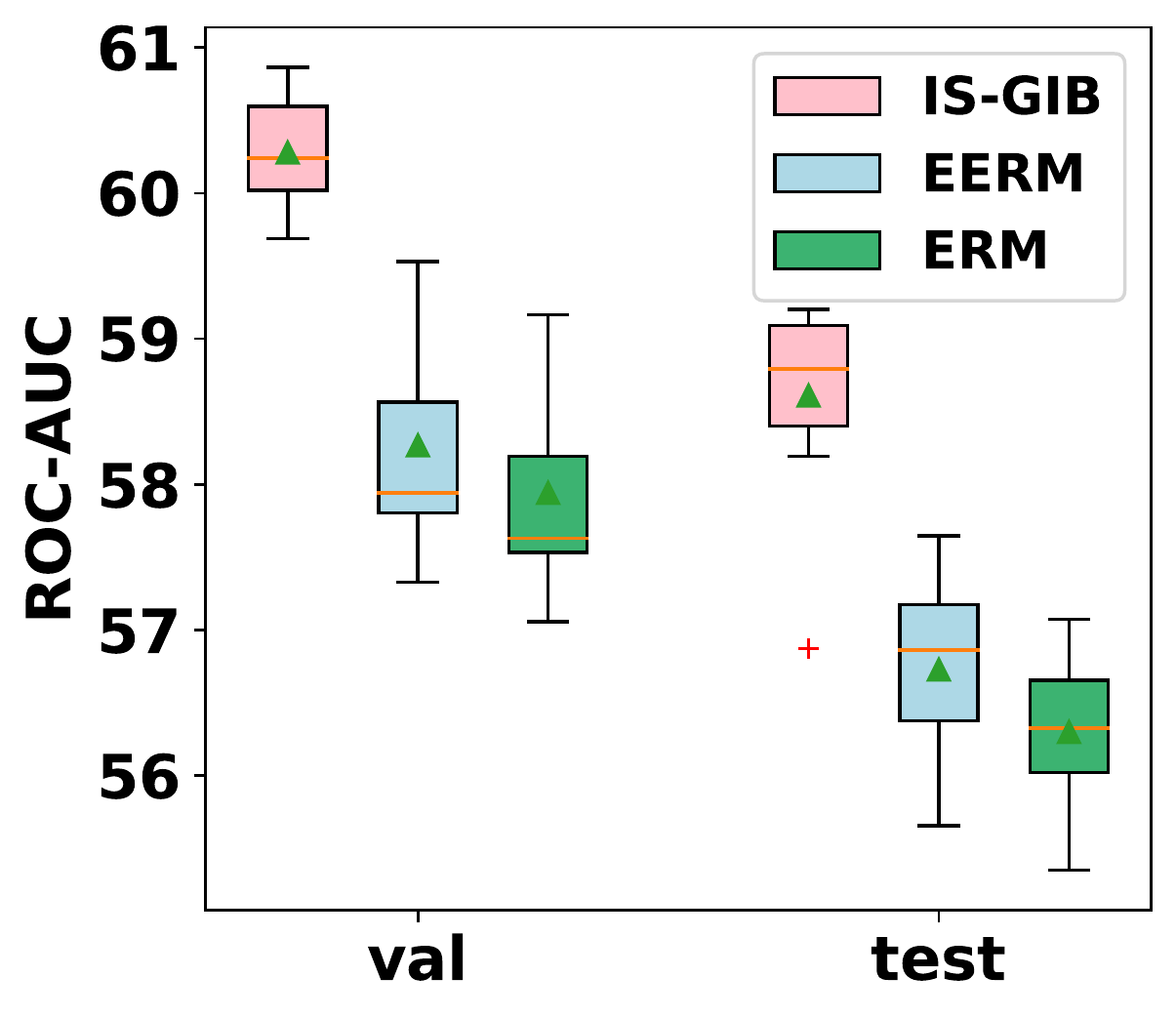}
  \end{minipage}
  }
  \subfigure[]{
  \begin{minipage}[t]{0.48\linewidth}
\makeatletter\def\@captype{figure}
    \centering
    \label{fig:42_1_2}
    \includegraphics[scale=0.32]{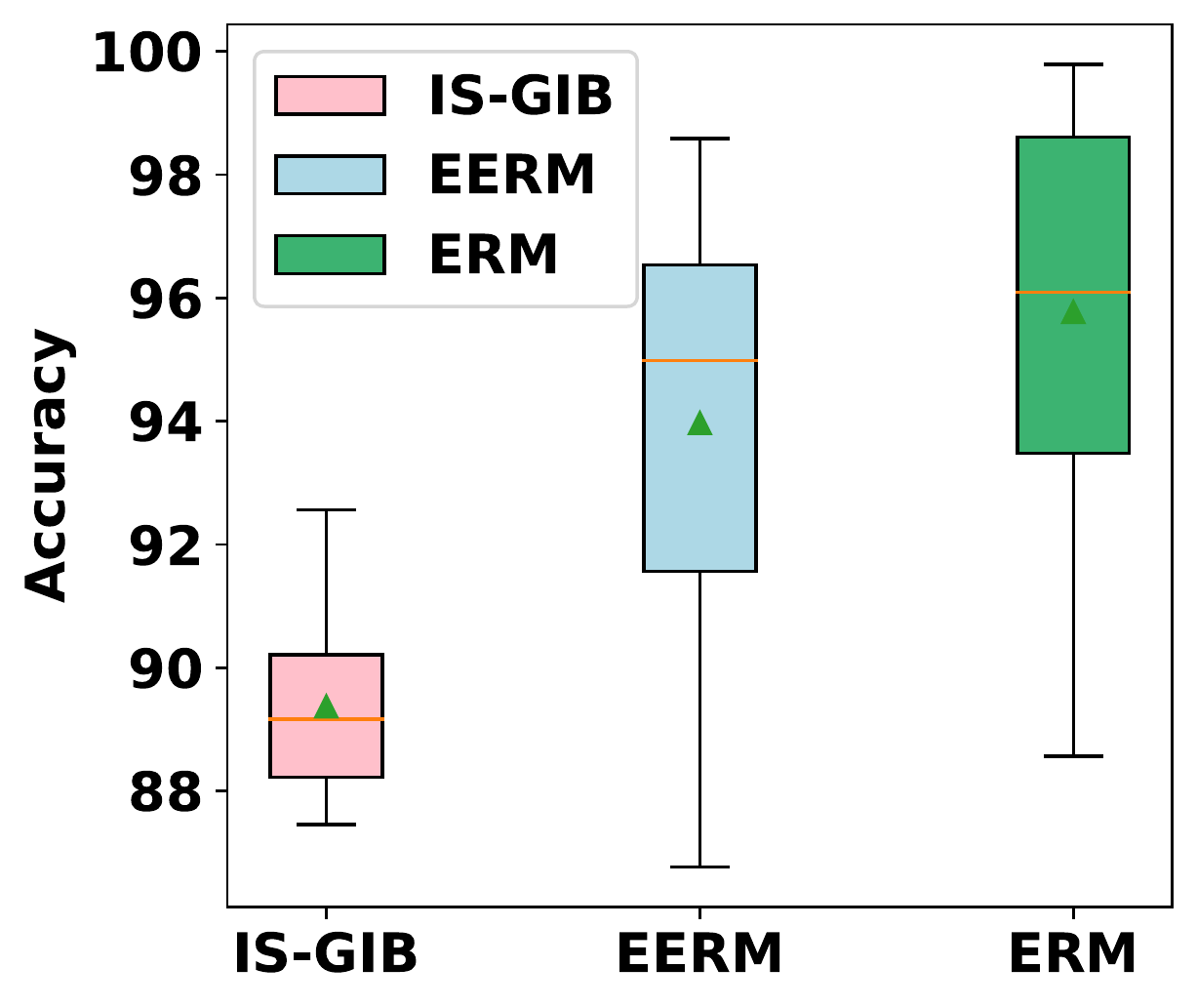}
  \end{minipage}%
  }
    \caption{Left: distribution of validation and test ROC-AUC. Right: distribution of training ROC-AUC.}
    \label{fig:42_1}
\end{figure}

\begin{figure}[h]
		\setlength{\abovecaptionskip}{0.2cm}
		    \centering
    \includegraphics[width=0.65\linewidth]{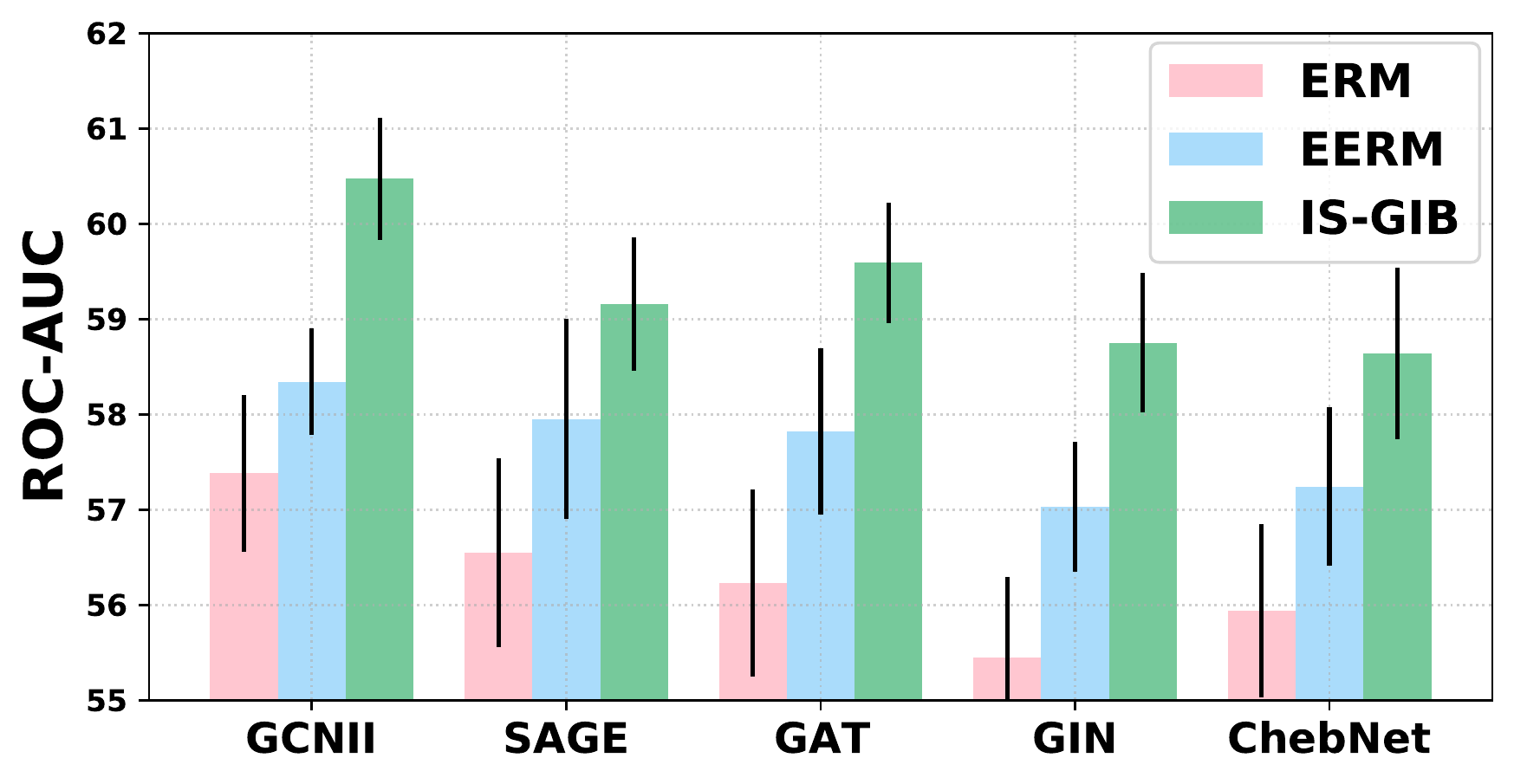}
		\caption{Test ROC-AUC of ERM, EERM, IS-GIB with different backbones.}
		 \label{fig:4_2_diff_backbone_boxplot}
\end{figure}

\noindent \textbf{Against Overfitting and Generalizing to Variant GNN Models.} Figure \ref{fig:42_1_1} shows the distribution of the inference results on testsets of IS-GIB, EERM and ERM with multiple runs on \textsc{Twitch-Explicit} dataset.
% (we use DE + FR for training, ENGB for validation, and the remaining graphs for testing). 
% We choose the model with the best performance on the development set as the final test model. 
We find: (1) The test performance of IS-GIB is much better than both EERM and ERM. (2) When achieving best performance on the development set, the training  ROC-AUC score of IS-GIB is the lowest among these three methods, which implies that our model can better avoid overfitting problem. We attribute such performance improvement o the explicit identification and suppression of spurious features by applying I-GIB and S-GIB.
% We attribute such generalization performance to the constraining of the mutual information between the input and the output of model.
Figure \ref{fig:4_2_diff_backbone_boxplot} shows the performance of IS-GIB using different GNN backbones on \textsc{Twitich-Explicit}.
% with the same configuration as the previous paragraph. 
For GCN, we use the newest version GCN-II (\cite{chen2020simple}). We find the GCN-II performs better than others, demonstrating that deeper GNN models can learn more useful domain-independent features with IS-GIB.
% across multiple graphs, thus combining IS-GIB with the GNNs which have better ability for tackling the over-smoothing is a better choice.

\subsection{Graph-Level OOD Graph Generalization}
In this subsection, 
we further focus on the graph-level classification task.
We choose two fake news detection datasets \textsc{Gossipcop} and \textsc{Politifact} collected by \cite{dou2021user}. Each of the graphs in these is a tree-structured fake/real news propagation graph extracted from Twitter. In each graph, the root node represents the news, the leaf nodes are Twitter users that retweet this news. The node feature is a 768-dimensional vector encoded by BERT. In this paper, we transform these directed tree-structured graphs into bi-direction graphs.
There are much more graphs than the former datasets, we randomly split all the graphs into training/validation/test sets for both of the two datasets. For \textsc{Gossipcop}, we randomly select 500 graphs for training, 200 graphs for validation, 300 graphs for testing. For \textsc{Politifact}, we randomly select 60 graphs for training, 30 graphs for validation, and using the other 224 graphs for testing. The selection is done by randomly shuffle the index three times with the random seed set to 2020, 2021, 2022 (corresponding to T1/T4, T2/T5, T3/T6 respectively).

\begin{table*}[ht]
\begin{center}
\begin{small}
\caption{Test accuracy on \textsc{Gossipcop} and \textsc{Politifact} with 3 different combination of training/validation/test graphs.}
% \vskip 0.15in
  \label{tab:5_3_graph_level}
  \center
  \setlength{\tabcolsep}{1.6mm}{
  \begin{tabular}{c|ccccccccc}
  \toprule
\multirow{2}{*}{Methods}  &
\multirow{2}{*}{\space}&
\multicolumn{3}{c}{Gossipcop} &
\multicolumn{1}{c}{\space}& \multicolumn{3}{c}{Politifact}&
\\ \cmidrule(r){3-5} \cmidrule(r){7-9} 
% \cline{9-10} 
&&T1 & T2 & T3&&T4& T5 & T6  \\
\midrule
ERM&&72.13$\pm$3.61&71.67$\pm$6.54&66.33$\pm$5.70&&70.85$\pm$3.75 &69.68$\pm$2.90&73.05$\pm$4.45 \\
\midrule
GroupDRO&&72.67$\pm$3.37&72.05$\pm$7.93&68.45$\pm$7.12&& 71.54$\pm$4.65&70.23$\pm$3.14&72.14$\pm$5.83\\
SR-GNN&&74.51$\pm$4.45&73.09$\pm$7.87&70.14$\pm$6.98&&70.98$\pm$3.12 & 70.36$\pm$4.49& 73.82$\pm$6.23 \\
OOD-GNN&&73.87$\pm$3.99&72.64$\pm$6.81&$69.44\pm$6.25&&71.14$\pm$3.68 & 70.95$\pm$3.26& 73.81$\pm$5.43 \\
\midrule
IRM&&72.98$\pm$4.86&72.12$\pm$6.33&68.86$\pm$7.09&&71.23$\pm$5.05 & 69.90$\pm$3.98& 73.91$\pm$5.15 \\
GOOD &&74.37$\pm$5.21&73.02$\pm$6.49&67.65$\pm$4.57&&71.20$\pm$6.03 & 70.64$\pm$5.27& 73.33$\pm$6.06 \\
Size-inv&&74.75$\pm$3.87&73.54$\pm$5.68&68.81$\pm$5.29&&72.50$\pm$4.64 & 72.31$\pm$6.07& 74.24$\pm$4.99 \\
DIR&&73.08$\pm$6.53&72.77$\pm$7.67&68.45$\pm$7.01&&70.94$\pm$5.43&70.68$\pm$5.71&73.47$\pm$4.89\\
EERM&&75.13$\pm$7.34&73.89$\pm$8.12&69.04$\pm$6.23&& 72.98$\pm$4.69&72.01$\pm$5.08&73.71$\pm$3.13 \\
\midrule
\textbf{IS-GIB}&&\textbf{78.47$\pm$4.96}&\textbf{76.60$\pm$6.99}&\textbf{71.24$\pm$5.08}&&\textbf{74.15$\pm$3.44}&\textbf{74.73$\pm$4.56}&\textbf{76.49$\pm$3.75}\\
\bottomrule
\end{tabular}}
\end{small}
\end{center}
\end{table*}

\begin{figure}[h]
%  \vspace{-0.2cm}  %调整图片与上文的垂直距离
		\setlength{\abovecaptionskip}{0.2cm}
% \setlength{\belowcaptionskip}{-0.1cm}   %调整图片标题与下文距离
% \vspace{-0.5em}
		\centering
% 		\vspace{-0.6cm}
		\includegraphics[width=0.65\linewidth]{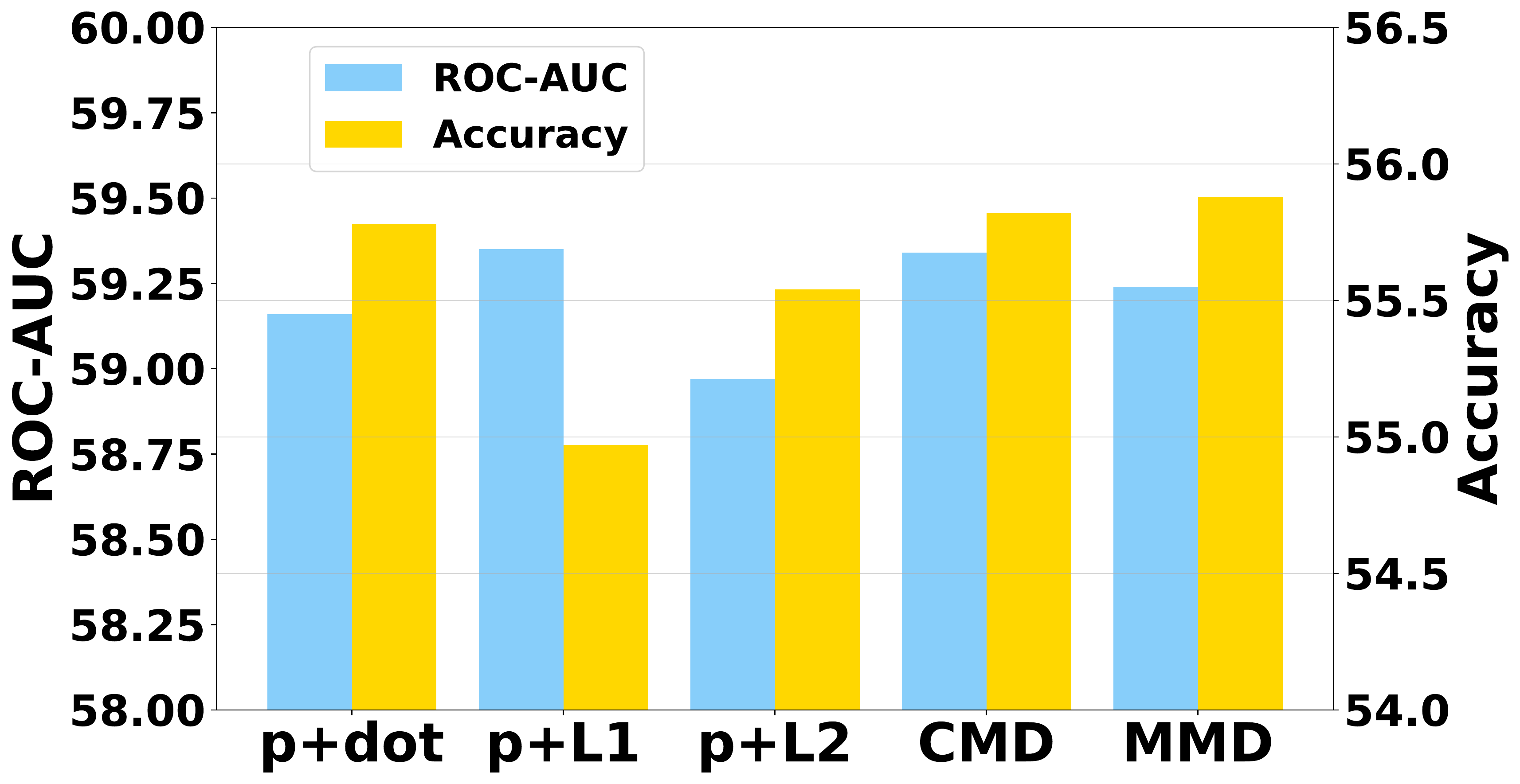}
		\caption{Test ROC-AUC (accuracy) on \textsc{TWitch-Explicit (Facebook100)} with different distance metrics.}
		\label{fig:5_1_ab2}
\end{figure}

Different from node-level classification problems, we need to add an extra readout function to the GNN model to get the final representation of the whole graph. 
From Table \ref{table_4_1_1} and Table \ref{tab:5_3_graph_level},
we observe that compared with node-level tasks, graph-level OOD generalization is a more challenging task. For graph-level classification, much more complex environments are introduced. While it is necessary to suppresses the effects of node-lvel fine-grained noise, we must deal with more graph-level coarse-grained noise.
% It can be seen that IS-GIB could also excel all other comparison baselines for the graph-level denoising.
% Compared with T1, T2 (T4, T5), the grpah selected by T3 and T6 are more diverse in scale, \cite{yehudai2021local} also mentioned distribution shifts caused by different size across graphs. 
We observe that IS-GIB still outperforms all of the baselines, which proves the effectiveness of the structural information bottleneck applied at graph level. And we also find that compared with Table \ref{table_4_1_1} and Table \ref{table_4_2_1_cross_d}, the deviation of experiment is much higher. We attribute this phenomenon to two aspects: (1) Graph-level OOD is much more complex. (2) Compared to the former two tasks, there is more randomness between different runs of experiments since we randomly select different combinations of training/validation/test graphs.\\

% In addition we find a similar experimental result to section \ref{sec_4_1}, when introducing more factors that lead to distribution drift, the performance gap between IS-GIB and baselines is amplified, demonstrating the superior of IS-GIB.

\begin{figure*}[h]
%  \vspace{-0.2cm}  %调整图片与上文的垂直距离
% \setlength{\belowcaptionskip}{-0.1cm}   %调整图片标题与下文距离
\centering
    \includegraphics[width=14cm]{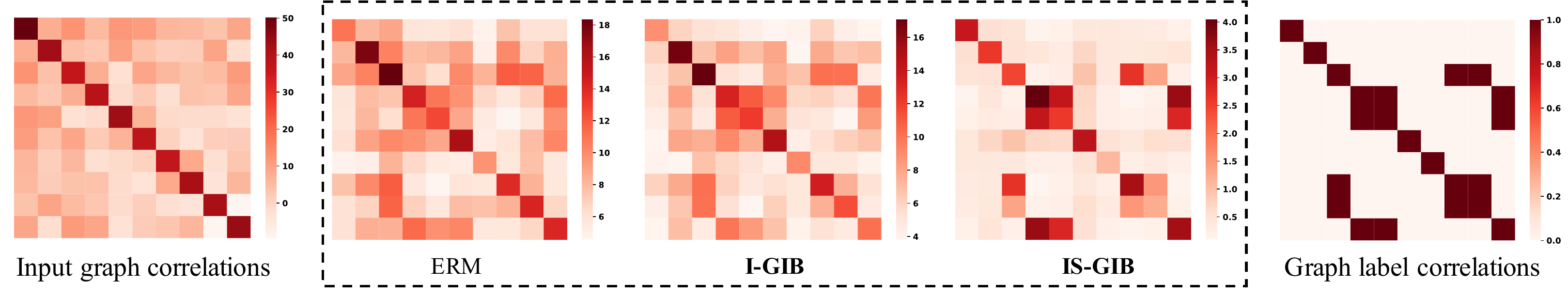}
\caption{Heat maps of pair-wise correlations. From left to right: input graph correlations, dashed box (baseline (ERM), I-GIB, IS-GIB), groundtruth graph label correlations. }
\label{fig:ablation_heatmap}
\end{figure*}

\noindent \textbf{Training Accuracy Comparison}. 
\begin{figure}[ht]
\vspace{-3mm}
%  \vspace{-0.2cm}  %调整图片与上文的垂直距离
%  \setlength{\abovecaptionskip}{0cm}   %调整图片标题与图距离
% \setlength{\belowcaptionskip}{-0.1cm}   %调整图片标题与下文距离
  \subfigure[]{
  
  \begin{minipage}[t]{0.48\linewidth}
\makeatletter\def\@captype{figure}
    \centering
    \label{fig:app_box1}
    \includegraphics[scale=0.32]{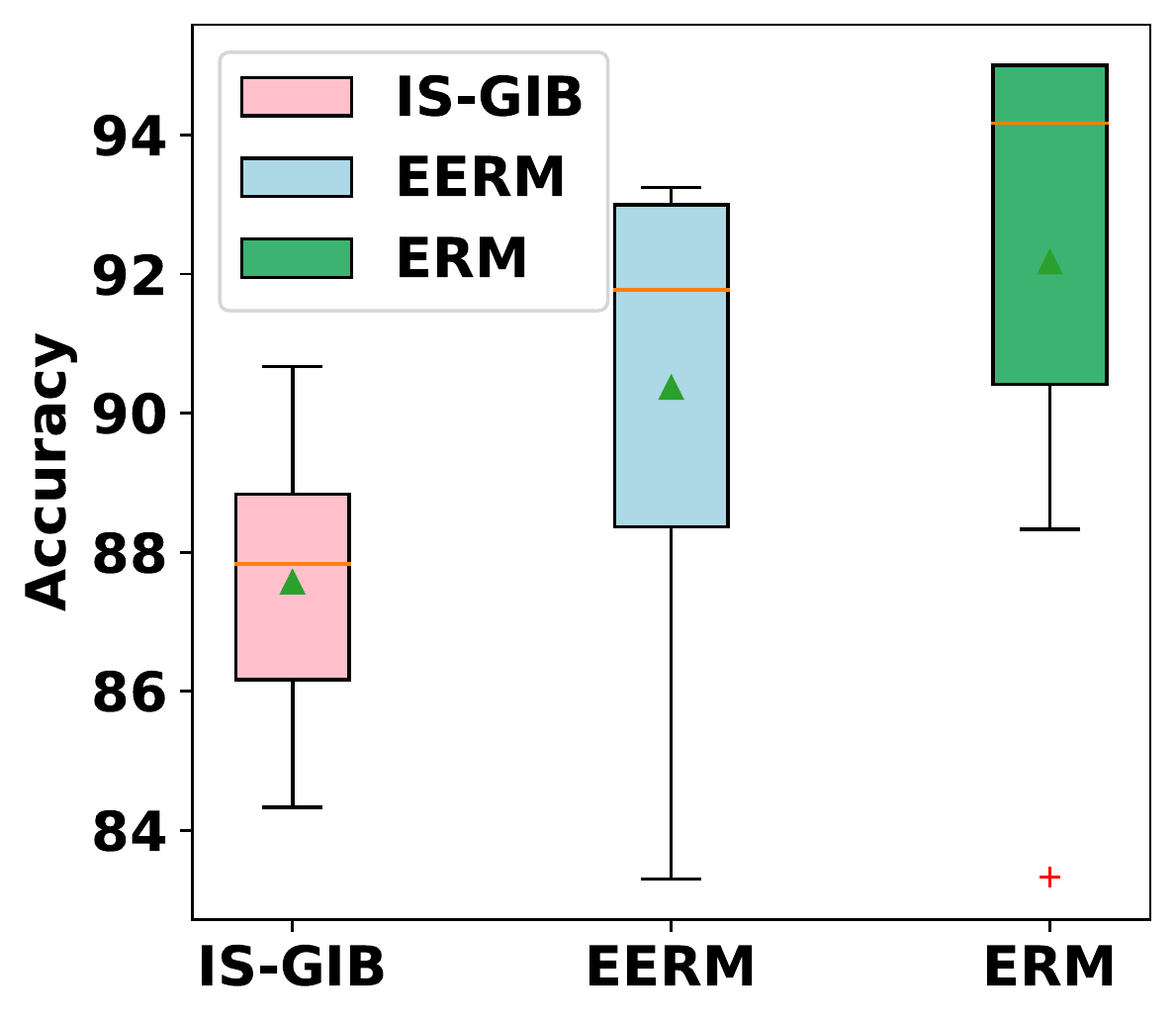}
    % \caption{train accuracy.}
  \end{minipage}
  }
  \subfigure[]{
  \begin{minipage}[t]{0.48\linewidth}
    \centering
    \label{fig:app_box2}
    \includegraphics[scale=0.32]{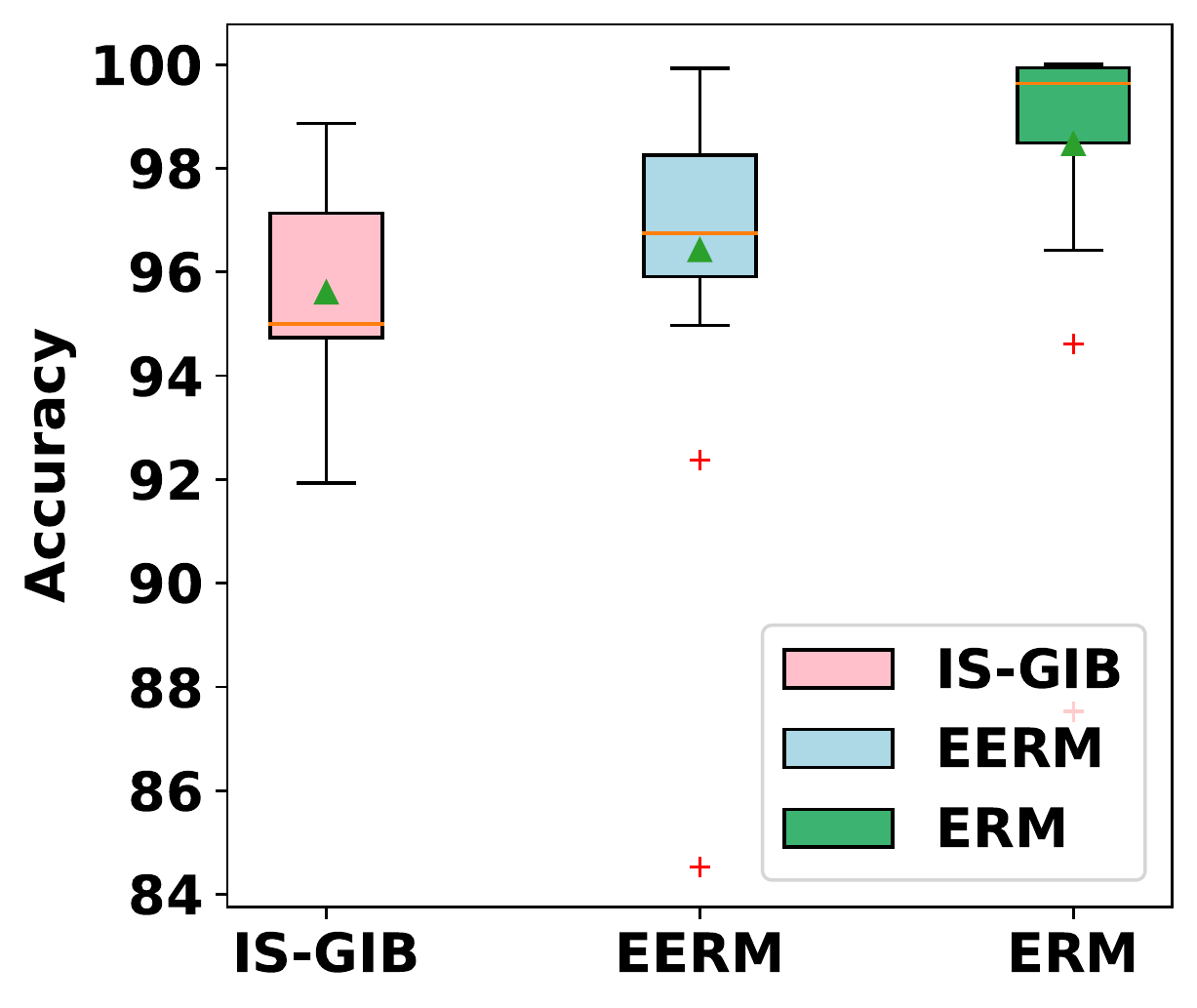}
  \end{minipage}%
  }
    \caption{(a) distribution of the mean value of training accuracy on \textsc{Gossipcop} and \textsc{Politifact}. (b) distribution of the mean value of training accuracy on \textsc{Cora} and \textsc{Citeseer}.}
    \label{fig:app_box}
    \vspace{-6mm}
    % \label{fig:dif_num_neighbors_per_hop}
\end{figure}
In section \ref{sec_4_2}, we have discussed the superior of IS-GIB in avoiding overfitting problems on \textsc{Twitch-Explicit}. It reaches the best validation performance with a lower training accuracy, which implies IS-GIB can better discard the useless information. In Figure \ref{fig:app_box}, we also show the distribution of training accuracy on \textsc{Cora}/\textsc{Citeseer} and \textsc{Gossipcop} and \textsc{Politifact} when reaching the highest accuracy on the development set. Again we find the deviation of training accuracy in graph-level tasks is larger than in node-level tasks,  and we can conclude that with a lower training accuracy, the model can get a better generalization performance than both the vanilla ERM and EERM in both graph-level and node-level tasks.

\section{Model Analysis}
\label{sec_ablation}
% Recall that the loss function contains four terms, among which the first term in Eq. \ref{eq11} is the vanilla ERM objective, thus we discuss the impact of weight of the other three terms. By default, we conduct this ablation experiment on \textsc{Twitch-Explicit} by adjusting one weight at a time and fixing others. The I-GIB min denotes the second term of Eq. \ref{eq4}, the S-GIB min (max)  denotes the second (first) term of Eq. \ref{eq7}.  Figure \ref{fig:5_1_ab1} shows the result, as we can see, a moderate value can result in a better performance. If the weight of the regular term is too large, the model will focus too much on generalization (i.e. similarity between domains) at the expense of more performance degradation on the testset. This performance degradation becomes more pronounced as the weights increase.

\textbf{Impact of I-GIB and S-GIB.}
Both Eq. \ref{eq10} and Eq. \ref{eq11} contains two terms: the first is the mutual information maximization (denoted by $\mathcal{L}_{*,1}$), the second is the mutual information restriction with coefficient $\beta$ (denoted by $\mathcal{L}_{*,2}$).
Note that $\mathcal{L}_{I,1}$ is the vanilla ERM objective, thus we first study the impact of the other three terms (See section \ref{sec:imple_detail} for the values of $\gamma_1, \gamma_2, \gamma_3$). Building on the vanilla ERM, we add other terms of the entire objective progressively, the results is illustrated in Table \ref{table_ablation_1}. As we can see, all of the terms have a positive impact on the model's performance. While $\mathcal{L}_{I,2}$ and $\mathcal{L}_{S,2}$ explicitly constrain the noise propagation on both instance level and inter-domain level, the $\mathcal{L}_{S,1}$ helps to learn the useful information between domains. In addition, the impact of $\mathcal{L}_{I,2}$ and $\mathcal{L}_{S,1}$ are more prominent than $\mathcal{L}_{S,2}$,
which indicates that it is vital to eliminate instance-level noises meanwhile preserving the invariant correlations between domains. I-GIB enables the model to learn more categorial information for better final predictions while the utilization of structural class relationships in S-GIB substantially improves the overall classification performance. 
\begin{center}
\begin{table*}[h]
\centering
\caption{Mean value of accuracy of different model variants on \textsc{Cora} \textsc{Facebook100} and \textsc{Gossipcop}.}
\label{table_ablation_1}
\begin{tabular}{c|cccc|cccc|cccc}
        \toprule
      & \multicolumn{4}{c}{\textsc{Cora}} &   \multicolumn{4}{c}{\textsc{Facebook100}} &   \multicolumn{4}{c}{\textsc{Gossipcop}} \\
      \midrule
    $\mathcal{L}_{I,1}$ &\checkmark &\checkmark  &\checkmark  &\checkmark   & \checkmark &\checkmark  &\checkmark  &\checkmark   & \checkmark &\checkmark&\checkmark&\checkmark\\
     $\mathcal{L}_{I,2}$ & -- &\checkmark  & \checkmark &\checkmark  & -- &\checkmark  & \checkmark &\checkmark  & -- &\checkmark&\checkmark&\checkmark\\
      $\mathcal{L}_{S,1}$ & --&--  & \checkmark & \checkmark & --&--  & \checkmark & \checkmark & --&--&\checkmark&\checkmark \\
       $\mathcal{L}_{S,2}$ & --& -- & -- &  \checkmark&  --& -- & -- &  \checkmark & -- &--&--&\checkmark\\
       \midrule
       Accuracy & 82.73& 85.74& 87.63& \textbf{89.15}& 72.13& 75.02& 77.39& \textbf{78.47} &  51.03& 53.68& 54.49& \textbf{55.24}\\
    % \checkmark & -- &-- &-- & 51.03 \\
    % \checkmark & \checkmark & --& --& 53.68 \\
    % \checkmark & \checkmark & \checkmark&--& 54.49 \\
    % \midrule
    % \checkmark & \checkmark & \checkmark&  \checkmark& \textbf{55.24} \\
    
    \bottomrule
    \end{tabular}
\end{table*}
\vspace{-6mm}
\end{center}

\noindent \textbf{Choices of Deterministic Relationship Measurements.}
\label{ablation-relationship-mea}
In Eq. \ref{eq10} (also Eq. \ref{eq7}), we compute the pairwise relationships $R(\cdot)$ among all the nodes (for node-level tasks) or graphs (for graph-level tasks). 
With respect to node-level task, 
we sample a subgraph surrounding the target node as the context feature, thus for both node- and graph-level tasks, we should exploit an extra readout function or a distance metrics on feature matrix.
% The time complexity of calculating this term is the square of the number of candidates. Thus for a more efficient optimization process, we turn to compute this regularization item by
% In addition, to achieve lower computational complexity, we sample a fixed batch of pairs for node-level tasks without sacrificing performance.
There are multiple choices of $R(\cdot)$, in the former experiments, we use mean-pooling with dot product (p+dot, and cosine similarity) by default. In this section, we explore more distance metrics: pooling + L1 norm (p+L1), pooling + L2 norm (p+L2),  CMD (Central moment discrepancy) \cite{zellinger2017central}, MMD (Maximum Mean Discrepancy) \cite{gretton2006kernel}. 
The formula for calculating CMD is:
\begin{equation}
\begin{aligned}
\label{eq_cmd}
CMD_K{[X,Y]}= & \frac{1}{|b-a|}||E(X)-E(Y)||_2 \\
& + \sum_{k=2}^n{\frac{1}{|b-a|^k}||C_k(X)-C_k(Y)||_2},
\end{aligned}
\end{equation}
Where $a, b$ are the bound of X and Y,  $E(X)=\frac{1}{|X|}\sum_{x\in X}x$ is
the empirical expectation vector of X, $C_k(X)=E[(X-E(X))^k]$  is the k-th order sample central moments of X. We set K=2.
The formula for calculating MMD is:
\begin{equation}
\label{eq_mmd}
MMD[f,X,Y]=||\frac{1}{n}\sum_{i=1}^n{f(X_i)}-\frac{1}{m}\sum_{j=1}^m{f(Y_j)} ||_\mathcal{H},
\end{equation}
Where $f$ maps a distribution to the Reproducing Hilbert Space. A kernel function is also needed for computing MMD, we choose Gaussian kernel and set the number of kernels to 3.
Figure \ref{fig:5_1_ab2} shows the performance of IS-GIB on \textsc{Twitch-Explicit} using different distance metrics. Compared with other metrics, CMD and MMD achieve a slight improvement in performance. 
Thus we conclude that it is important to consider the pairwise relationships between instances.It is better to take fine-grained structural information into consideration for the correlative distance matching.

\noindent \textbf{Structural Denoising with IS-GIB.}
We further interpret what is of importance the S-GIB learns in practice. Visualization results are shown in Figure.\ref{fig:ablation_heatmap}, we draw heat maps of three parts: the correlations of input graphs, the final representations (in the dashed box), and graph labels. Note that compared with the baseline ERM and our I-GIB, the IS-GIB can significantly boost the structural denoising from the input to the representations, its heat map is structurally sharper supported by the proposed S-GIB and encourages more accurate predictions.

\begin{figure}[h]
% \vspace{-2mm}
\centering
    \includegraphics[width=5cm]{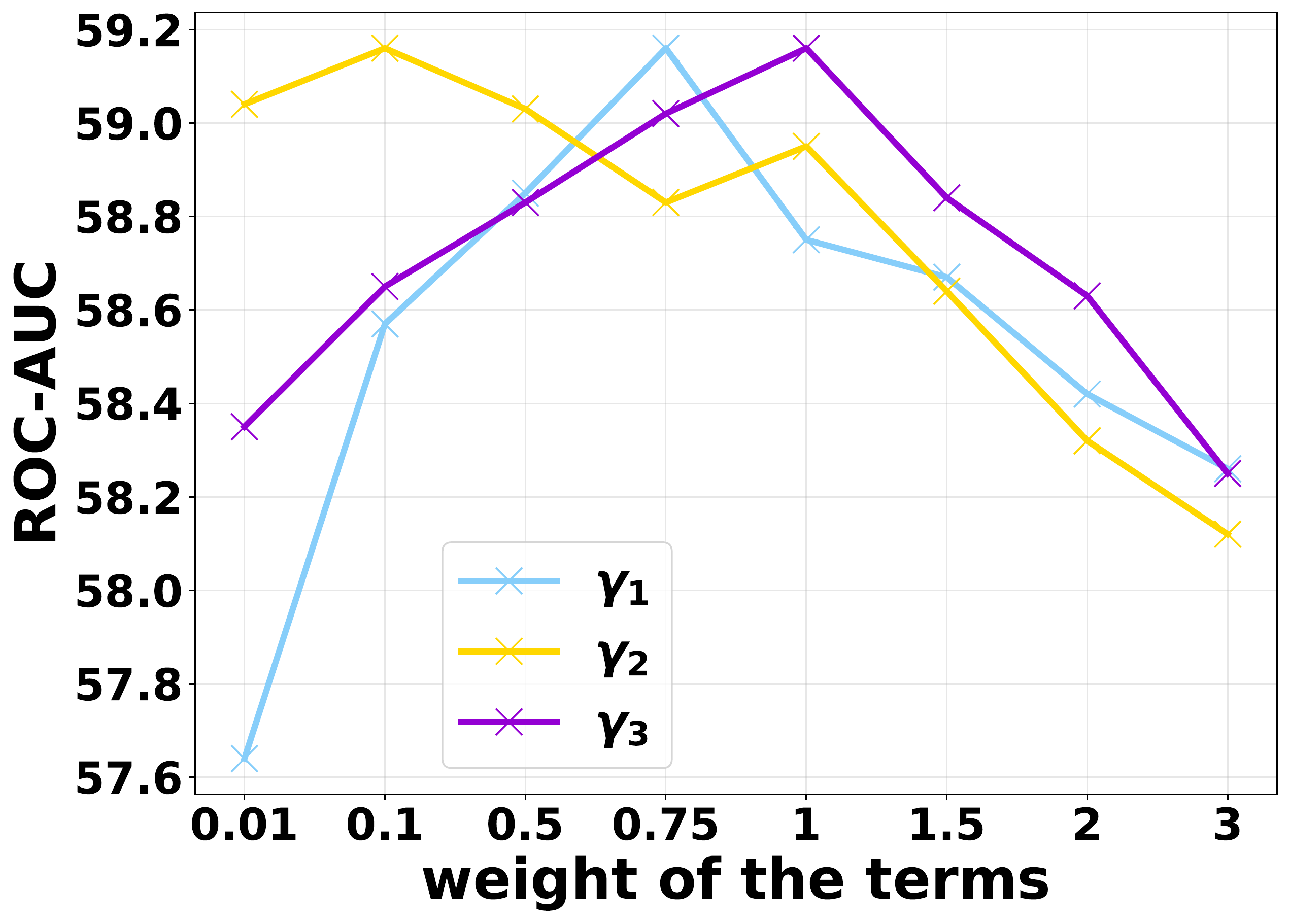}
% \vspace{-3mm}
\caption{Test ROC-AUC score on \textsc{Twitch-Explicit} with different regularization weights.}
\label{fig:appendix_diff_weight}
% \vspace{-2mm}
\end{figure}
\noindent\textbf{Weights Influence of Optimization Objective.}
Recall that in the objective Eq. \ref{eq12}, the loss function contains four terms. In order to better fit the training data, we need to pay attention to the ERM item; in order to achieve better generalization performance, we also need to pay attention to different information bottleneck items.
% with different order of magnitude, 
thus it is necessary to consider their respective different weights to achieve better generalization performance while fully fitting the training data. 
% Similar to section \ref{sec_ablation},
% % to study the impact of the weights of different parts of Eq. \ref{eq12},
% we rewrite the objective Eq. \ref{eq12}:
Recall the final objective is:
\begin{equation}   \mathcal{L}_{total}=\mathcal{L}_{I,1}+\gamma_1\mathcal{L}_{I,2}+\gamma_2\mathcal{L}_{S,1}+\gamma_3\mathcal{L}_{S,2}
\end{equation}
% The weight of three different regularization terms is also crucial to the performance. 
% Thus in this section we discuss the impact of weight of them.
We conduct experiments with different values of $\gamma_1, \gamma_2, \gamma_3$.
By default, we conduct this ablation experiment on \textsc{Twitch-Explicit} by adjusting one weight at a time and fixing others. 
First, we searched for parameter combinations in the value range \{1e-3, 1e-2, 1e-1, 1, 5, 10\}, identified better performing sub-ranges, and then performed fine-grained search between 1e-2 and 3.
% The I-GIB min denotes the second term of Eq. \ref{eq4}, the S-GIB min (max)  denotes the second (first) term of Eq. \ref{eq7}. 
Figure \ref{fig:appendix_diff_weight} shows the result, 
% as we can see, different regularization terms have different magnitudes, finding a moderate value for each of them can result in a better performance. If the weight of the regular term is too large, the model will focus too much on generalization (i.e. similarity between domains) at the expense of more performance degradation on the testset. This performance degradation becomes more pronounced as the weights of the terms increase.
as we can see, different information bottleneck items are roughly in the same order of magnitude. We need to find appropriate values for different terms respectively to make the model perform the best. On one hand if the weight is too small, the performance of the model will drop a little bit, which means that the noise removal and cross domain invariant feature learning is insufficient, which is close to the performance of standard ERM. On the other hand, assigning a large weight to the IB terms is also not a good choice. Obviously, the optimization of the IB term will be more difficult than the ERM term. Excessive enlargement of the weight of the IB term will make the model unable to fully optimize the ERM term, which will lead to underfitting.

\noindent\textbf{Time Complexity.}
In this paper, we propose Individual and Structural Graph Information Bottlenecks (IS-GIB) for graph invariant learning. Let $n$ be the number of nodes in training set, d be the hidden size of GNN, $\Phi (n,d)$ be the time that GNN takes to compute node embedding of the whole graph. The first item in equation (\ref{eq11}) takes extra $O(n)$ time, and the second item takes $O(b\cdot n \cdot d)$ and $b$ is the negative sampling size. Let $R(d)$ be the time for computing the deterministic relationship with embedding size $d$. The first item in equation (\ref{eq10}) which compute the pair-wise relationship takes $O(n^2\cdot R(d))$ time, and the second one takes $O(b\cdot n^2\cdot R(d))$ time. Thus the total complexity of IS-GIB is $O(\Phi (n,d)+b n^2 \cdot max\{R(d),d\})$.

% \subsection{Training Accuracy Comparison}

\section{Conclusion}
In this paper, we endeavor to handling the problem of Out-of-Distribution (OOD) graph generalization. We propose a complementary framework Individual and Structural Graph Information Bottlenecks (IS-GIB) to conduct graph invariant learning. The Individual Graph Information Bottleneck (I-GIB) imposes explicit instance-wise constraints for discarding spurious feature in node- or graph-level representations while Structural Graph Information Bottlenecks (S-GIB) focuses on exploiting pair-wise intra- and inter-domain correlations between node or graph instances. 
We theoretically derive a tractable optimization objective for IS-GIB and consistently achieve the best performance on extensive graph datasets across domains. For future work, we would explore to extend the algorithm to other more challenging graph-based application problems \cite{yang2022diffusion}.

% or



% \ifCLASSOPTIONcompsoc
%   % The Computer Society usually uses the plural form
%   \section*{Acknowledgments}
% \else
%   % regular IEEE prefers the singular form
% \fi


\ifCLASSOPTIONcaptionsoff
  \newpage
\fi



%\bibliographystyle{IEEEtran}


% \end{thebibliography}
\bibliographystyle{IEEEtran}
\bibliography{references.bib}

\begin{IEEEbiography}[{\includegraphics[width=1in,height=1in,clip,keepaspectratio]{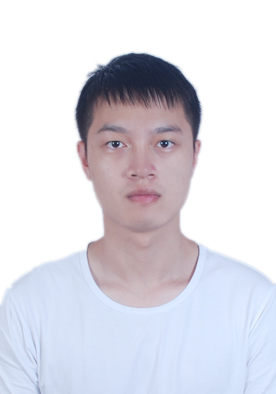}}]{Ling Yang} is currently a Ph.D. student in Peking University (PKU), Beijing, China. He is working in National Institute of Health Data Science of Peking University. His research interests include graph machine learning and representation learning. He serves as a reviewer for TPAMI, NeurIPS, ICML, CVPR, KDD, AAAI.
\end{IEEEbiography}

\begin{IEEEbiography}[{\includegraphics[width=0.8in,height=1in,clip,keepaspectratio]{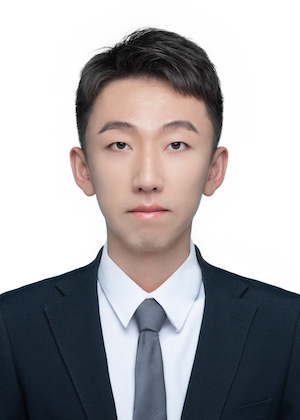}}]{Jiayi Zheng}
received his B.S. and M.S. from the School of Computer Science, Peking University (PKU), Beijing, China. His research interests include graph neural networks, natural language understanding, and recommender systems. He has published several papers in the top conferences including IJCAI, ICLR etc.
\end{IEEEbiography}

\begin{IEEEbiography}[{\includegraphics[width=0.8in,height=1in,clip,keepaspectratio]{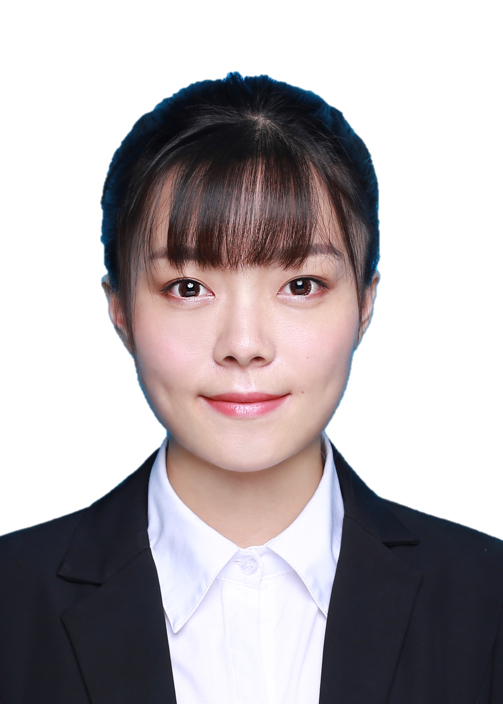}}]{Heyuan Wang} 
received her Ph.D. degree from School of Computer Science, Peking University (PKU), in 2023. Her research interests include spatiotemporal modeling, recommender system and natural language understanding. She has published several papers in TKDE, IJCAI, AAAI, ACL and other international academic conferences and journals.
\end{IEEEbiography}

\begin{IEEEbiography}[{\includegraphics[width=0.8in,height=1in,clip,keepaspectratio]{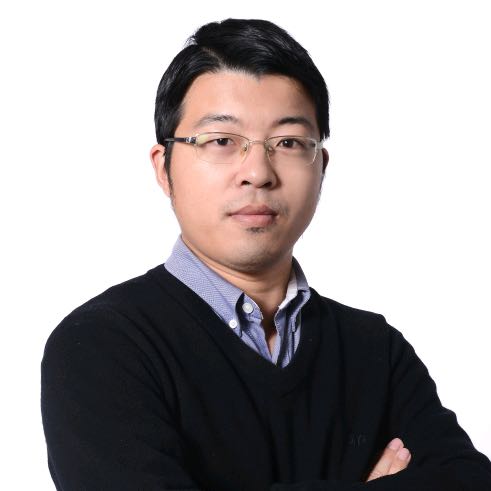}}]{Zhongyi Liu}
received his B.S. and M.S. degrees in computer science from Peking University, China. Currently, he is a technical director of Ant Group. He is responsible for the search and recommendation technology
team in Ant Group and his research interests lie in recommender systems, information retrieval, machine learning and related domains. Before joining Ant Group in 2017, he was a technical director of Alibaba Group.
\end{IEEEbiography}

\begin{IEEEbiography}[{\includegraphics[width=0.8in,height=1in,clip,keepaspectratio]{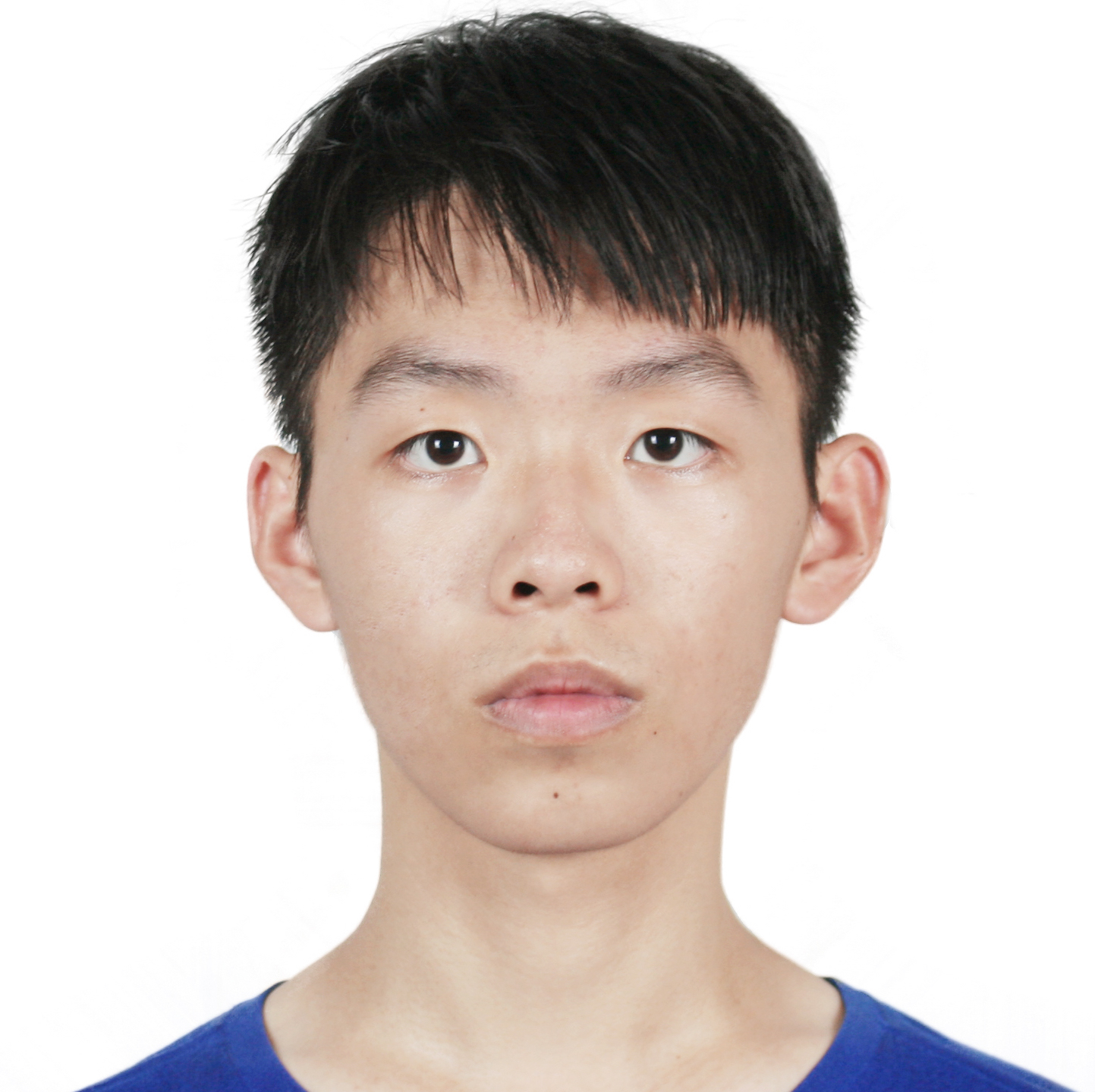}}]{Zhilin Huang}
received his B.S. from the School of Electronics and Information Engineering, Harbin Institute of Technology in 2019. He received his M.S. from the School of Electronic and Computer Engineering, Peking University in 2022. His research interests include computer vision, self-supervised learning, computational biology.
\end{IEEEbiography}

\begin{IEEEbiography}[{\includegraphics[width=0.8in,height=1in,clip,keepaspectratio]{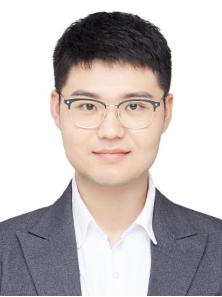}}]{Shenda Hong}
is an Assistant Professor in National Institute of Health Data Science at Peking University. His research interests are data mining and artificial intelligence for real-world healthcare data, especially deep learning for temporal medical data – such as temporal events, time series (e.g. longitudinal data, electronic health records, claims data), and physiological signals (e.g. ECG, EEG, PSG). He serves as an associate editor of Health Data Science - a Science Partner Journal.
\end{IEEEbiography}

\begin{IEEEbiography}[{\includegraphics[width=0.8in,height=1in,clip,keepaspectratio]{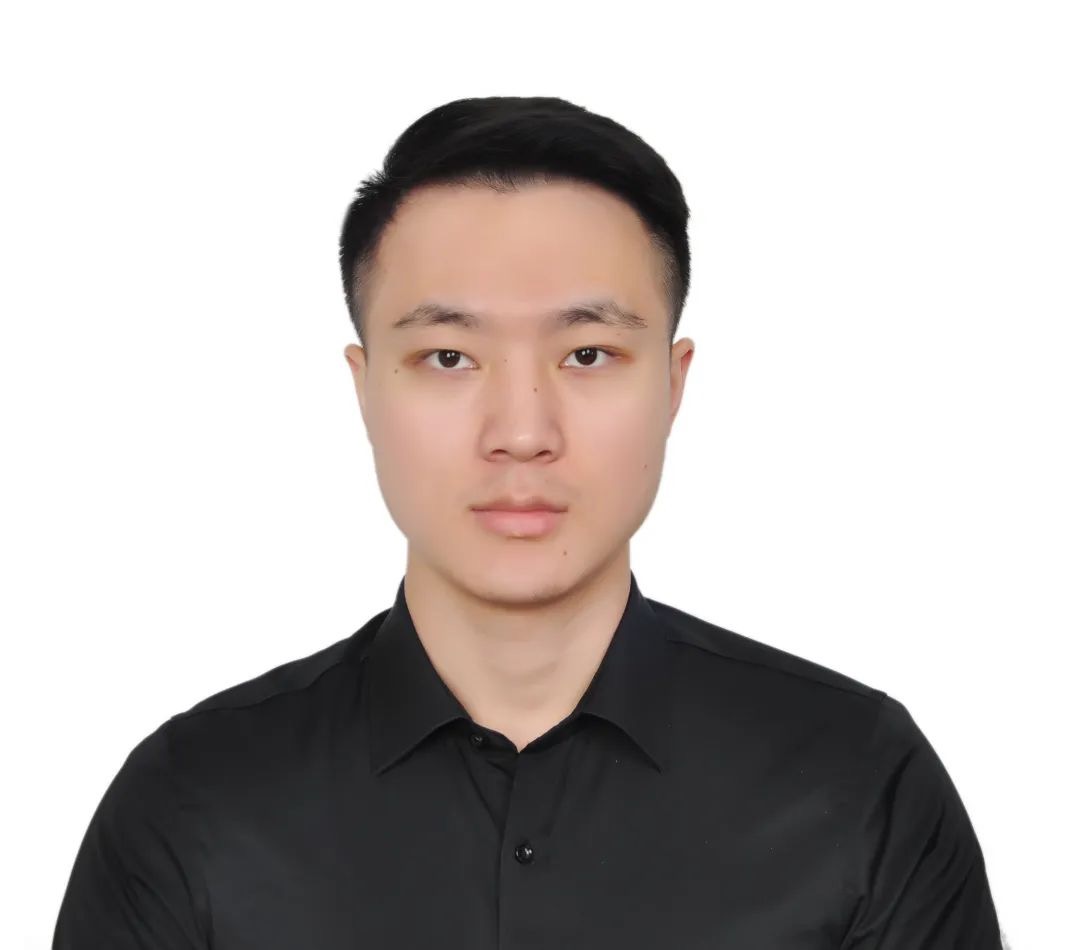}}]{Wentao Zhang}
is currently a postdoc research fellow at Montreal Institute for Learning Algorithms (Mila). Wentao’s research focuses on large-scale graph learning from three perspectives – data, model, and system. His research works have been powering several billion-scale applications in Tencent, and some of them have been recognized by multiple prestigious awards, including the Best Student Paper Award at WWW’22.
\end{IEEEbiography}

\begin{IEEEbiography}[{\includegraphics[width=0.8in,height=1in,clip,keepaspectratio]{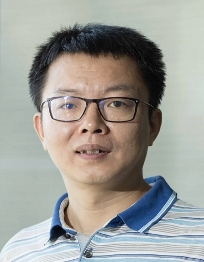}}]{Bin Cui}
is a professor in the School of EECS and
Director of Institute of Network Computing and
Information Systems, at Peking University. His
research interests include database systems,
and data mining. Prof. Cui has published more
than 100 research papers, and is the winner
of Microsoft Young Professorship award (MSRA
2008), and CCF Young Scientist award (2009).
\end{IEEEbiography}


%\newpage


%\vfill


\end{document}